\documentclass[a4paper,fleqn]{cas-dc}
\usepackage[numbers]{natbib}
\usepackage{algorithm}
\usepackage{algorithmic}

\begin{document}
\let\WriteBookmarks\relax
\def\floatpagepagefraction{1}
\def\textpagefraction{.001}

\shorttitle{EndoCogniAgent: Closed-Loop Agentic Reasoning for Endoscopic Diagnosis}

\shortauthors{Y. Tang et~al.}

\title [mode = title]{EndoCogniAgent: Closed-Loop Agentic Reasoning with Self-Consistency Validation for Endoscopic Diagnosis}

\author[1]{Yi Tang}
\fnmark[1]
\ead{tangyi_1218@seu.edu.cn}
\credit{Conceptualization, Data curation, Investigation, Methodology, Software, Validation, Visualization, Writing -- original draft}

\author[2]{Kai-Ni Wang}
\fnmark[1]
\ead{kainiwang@cuhk.edu.hk}
\credit{Data curation, Investigation, Supervision, Validation, Writing -- review and editing}

\author[4]{Liang-Peng Pu}
\ead{puliangpengnjmu@163.com}
\credit{Data curation, Investigation, Validation, Writing -- review and editing}

\author[3]{Hui Tang}
\cormark[1]
\ead{corinna@seu.edu.cn}
\credit{Funding acquisition, Project administration, Resources, Supervision}

\author[4]{Xiaopu He}
\cormark[1]
\ead{help_007@126.com}
\credit{Data curation, Investigation, Validation, Writing -- review and editing}

\author[1]{Guang-Quan Zhou}
\cormark[1]
\ead{guangquan.zhou@seu.edu.cn}
\credit{Conceptualization, Funding acquisition, Project administration, Resources, Supervision, Writing -- review and editing}

\cortext[1]{Corresponding author.}
\fntext[1]{These authors contributed equally to this work.}

\affiliation[1]{organization={School of Biological Science and Medical Engineering, Southeast University},
            city={Nanjing},
            postcode={211189},
            country={China}}

\affiliation[2]{organization={Department of Computer Science and Engineering, The Chinese University of Hong Kong},
            city={Hong Kong},
            country={China}}

\affiliation[3]{organization={School of Computer Science and Engineering, Southeast University},
            city={Nanjing},
            postcode={210096},
            country={China}}

\affiliation[4]{organization={The First Affiliated Hospital of Nanjing Medical University},
            city={Nanjing},
            country={China}}

\begin{abstract}
Endoscopic diagnosis is an iterative process in which clinicians acquire, compare, and verify local visual evidence before reaching a conclusion. Current AI systems do not adequately support this process because fine-grained evidence acquisition and multi-step reasoning remain weakly coupled, complicating reconciliation of image-derived findings with their textual interpretations. This gives rise to two failure modes, hallucinated evidence and uncorrected error accumulation, that undermine diagnostic reliability. We propose EndoCogniAgent, a closed-loop agentic framework that formulates endoscopic diagnosis as a controlled state update process for integrating complementary visual and textual evidence. At each reasoning round, a central planner selects an evidence acquisition action, specialized expert tools extract spatial and semantic observations as structured textual evidence, and a self-consistency validation mechanism examines this evidence along two dimensions, knowledge consistency against the input image and temporal consistency with prior validated findings, before updating the diagnostic state. Validated observations are admitted into the evolving state to condition subsequent planning, while insufficiently supported or conflicting findings are retained with corrective feedback that redirects the planner toward additional verification.
We further introduce EndoAgentBench, a workflow-oriented benchmark comprising 6,132 question-answer pairs from 11 endoscopic datasets, designed to evaluate diagnostic agents across a comprehensive diagnostic chain, from fine-grained visual perception to high-level diagnostic reasoning. Experiments show that EndoCogniAgent achieves 85.23\% overall accuracy on perception tasks and a 71.13\% clinical acceptance rate on reasoning tasks. Blinded clinician evaluation further shows consistent improvements in diagnostic response quality over the evaluated baselines.
Code and data are available at \url{https://github.com/Tyyds-ai/EndoCogniAgent}.
\end{abstract}


\maketitle

\section{Introduction}
Endoscopic diagnosis relies on iterative evidence accumulation because clinically decisive cues are often local and mutually dependent. Clinicians rarely reach a conclusion from a single visual sign. Instead, they gather, compare, and cross-check multiple findings before forming a stable judgment. For example, lesion characterization often depends on the joint assessment of surface pattern, vascular structure, and boundary morphology, where no single cue is sufficient on its own~\cite{cassinotti2023endoscopic}. This poses a challenge for current AI systems, which must not only capture fine-grained evidence at the relevant spatial scale, but also maintain diagnostic coherence as new findings are incorporated over time~\cite{madeline2026computer}.

\begin{figure}
    \centering
    \includegraphics[width=0.9\linewidth]{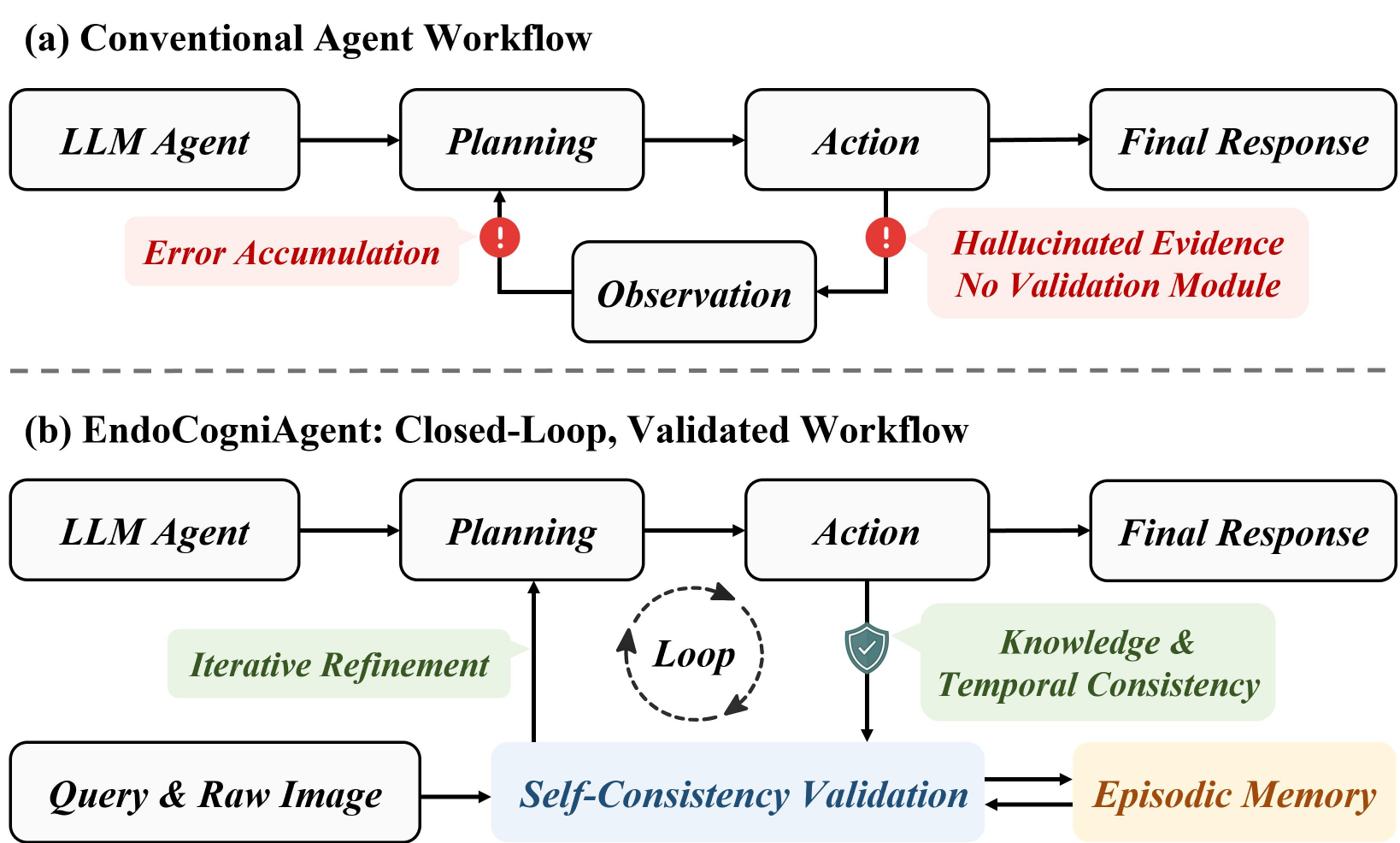}
    \caption{Workflow Comparison. EndoCogniAgent (b) vets observations via closed-loop validation, mitigating hallucinations and error accumulation found in uncontrolled pipelines (a).}
    \label{fig:contrast}
\end{figure}

\begin{figure*}[t]
    \centering
    \includegraphics[width=0.95\textwidth]{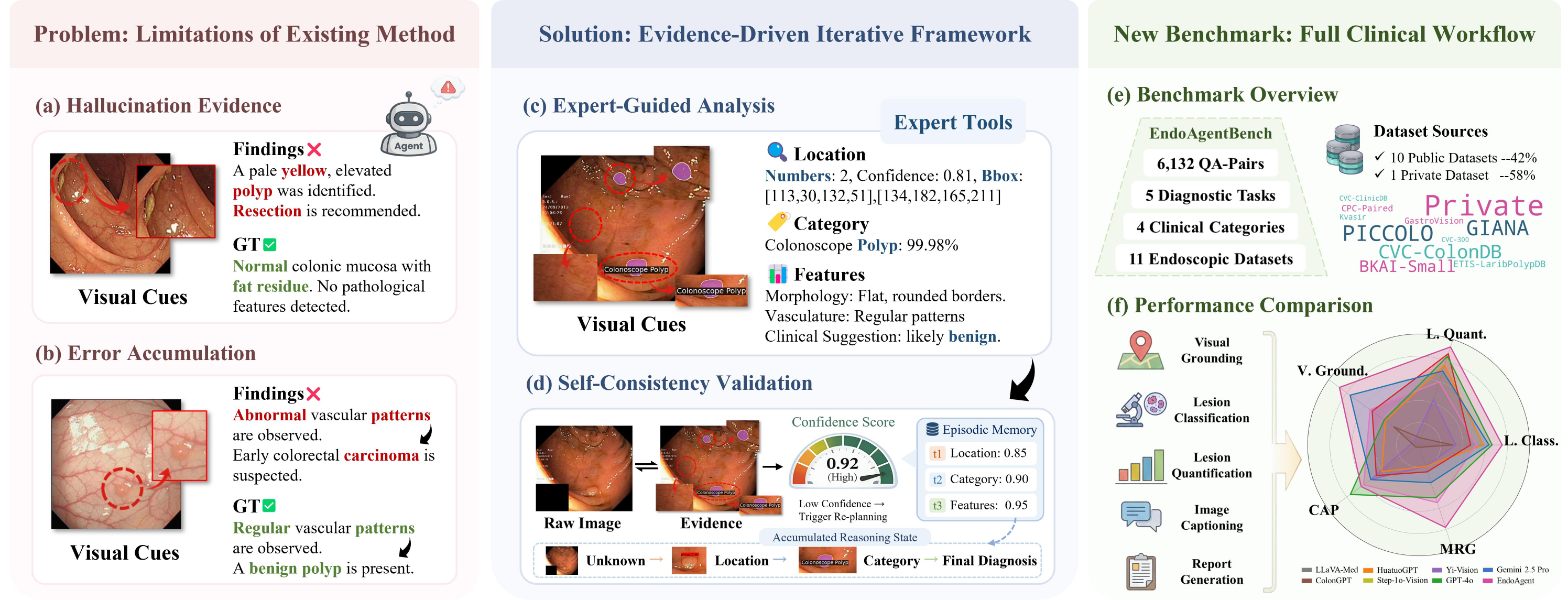}
    \caption{Overview of EndoCogniAgent and EndoAgentBench. (a-b) illustrate two key failure modes of existing multimodal models in endoscopic diagnosis, namely hallucinated evidence and error accumulation. (c-d) present the proposed closed-loop framework, in which expert evidence extraction is followed by self-consistency validation before newly acquired evidence can update the diagnostic state. (e) shows EndoAgentBench, a workflow-oriented benchmark comprising 6,132 question-answer pairs across five diagnostic tasks from 11 endoscopic datasets. (f) compares performance on fine-grained visual perception and high-level diagnostic reasoning tasks.}
    \label{fig:overview}
\end{figure*}

Current endoscopic AI models do not fully support this complex form of diagnosis because fine-grained perception and coherent diagnostic reasoning are not effectively integrated~\cite{bencardino2025artificial}. Task-specific models can localize lesions or predict pathology with strong performance, but they usually operate in isolation and do not organize heterogeneous findings into an evolving diagnostic process. Multimodal large language models (MLLMs) offer a more flexible interface for question answering and report generation, yet their visual grounding is often less reliable in subtle endoscopic scenes, where small local cues can determine the final judgment~\cite{liu2026medical}. Across both lines of work, a key challenge is to integrate image-derived findings with textual interpretations while checking whether they are visually supported and mutually consistent before they guide later decisions.

Agent-based reasoning provides a promising framework for endoscopic diagnosis, but existing agents still do not adequately prevent \textbf{hallucinated evidence} and \textbf{error accumulation}~\cite{liu2025medmmv}. Recent medical agents have shown the value of this design by decomposing inference into planning and tool execution. However, their limitation is not merely incomplete verification, but the lack of explicit control over how intermediate evidence enters and reshapes the reasoning state~\cite{cemri2025multi}. As illustrated in Fig.~\ref{fig:overview}a, hallucinated evidence arises when observations inferred from weak or ambiguous local cues are accepted as valid evidence before sufficient image support has been established. Similarly, error accumulation (Fig.~\ref{fig:overview}b) occurs when early findings are not re-evaluated as later evidence becomes available, allowing local inconsistencies to propagate through the reasoning chain. These failure modes are inherent in conventional uncontrolled pipelines (Fig.~\ref{fig:contrast}a), where small early errors can alter the entire downstream diagnostic path.

We address these limitations with \textbf{EndoCogniAgent}, a closed-loop framework that formulates endoscopic diagnosis as a controlled state update process (Fig.~\ref{fig:contrast}b). To support trustworthy endoscopic diagnosis, our design integrates visual evidence and textual interpretations through consistency checking, so that unsupported or conflicting findings can be reconsidered during subsequent reasoning.
Specifically, a Central Decision Unit (CDU) plans which type of clinical evidence should be acquired next, and a Perceptual Expert Toolset (PET) provides the corresponding fine-grained observations through specialized models for core endoscopic tasks (Fig.~\ref{fig:overview}c).
These observations, including spatial predictions and semantic descriptions, are converted into structured textual evidence for joint reasoning, while the original image provides the visual basis for checking their support.
This decoupling improves evidence acquisition, but it also makes explicit control over state update essential, because heterogeneous tool outputs may otherwise introduce weakly supported findings or unresolved conflicts into later reasoning. EndoCogniAgent therefore introduces an internal self-consistency validation mechanism (ISCV) that examines each new observation before it is incorporated into subsequent decisions (Fig.~\ref{fig:overview}d). The validation operates along two complementary dimensions. Knowledge consistency checks whether an observation is grounded in visible image cues, while temporal consistency checks whether it remains coherent with prior findings across reasoning rounds. The resulting feedback is then written into the episodic memory module (EMM) and used to condition subsequent planning on validated intermediate states rather than raw tool outputs. In this way, the framework links visual--textual evidence integration with explicit conflict checking, providing a mechanism for reconsidering unsupported or inconsistent findings during diagnostic reasoning.

Evaluation should follow the same workflow-level view of endoscopic diagnosis. We therefore construct \textbf{EndoAgentBench}, a benchmark with 6,132 visual question-answer pairs from 11 endoscopic datasets, designed to evaluate diagnostic agents across a comprehensive diagnostic chain rather than isolated model capabilities alone (Fig.~\ref{fig:overview}e). The benchmark unifies fine-grained visual perception to high-level diagnostic reasoning, bridging the gap between local evidence acquisition and comprehensive report synthesis within a consistent evaluation framework. To support clinically meaningful assessment of report-level outputs, we further introduce Clinical Acceptance Rate (CAR) through an LLM-as-a-Judge paradigm. Experiments confirm that closed-loop validation yields superior performance across the diagnostic workflow (Fig.~\ref{fig:overview}f).
This work makes four main contributions:

\begin{itemize}
\item  To the best of our knowledge, we present the first endoscopy-specific agent framework for closed-loop diagnostic reasoning, namely EndoCogniAgent, which formulates endoscopic diagnosis as an iterative process of planning, expert evidence extraction, and validation guided state update.
\item  We develop an ISCV mechanism that checks the visual support and cross-step coherence of textual evidence, guiding the integration of complementary observations and the reconsideration of unsupported or conflicting findings during multi-step reasoning.
\item We introduce EndoAgentBench, a comprehensive workflow-oriented benchmark comprising 6,132 visual question-answer pairs from 11 endoscopic datasets, to support unified evaluation of fine-grained visual perception and high-level diagnostic reasoning for endoscopic diagnostic agents.
\item Extensive experiments show that EndoCogniAgent delivers clear gains over general-purpose, medical, and agentic baselines, achieving 85.23\% average accuracy on visual tasks and 71.13\% CAR on language tasks, with ablation analysis confirming that ISCV and EMM are individually critical to these gains.
\end{itemize}

\section{Related Works}

\subsection{Endoscopic Foundation Models and Medical MLLMs}
Recent endoscopic AI research mainly advances along two directions: specialized perceptual models and medical MLLMs. Foundation-style models such as EndoDINO and EndoMamba learn transferable representations from large-scale endoscopic corpora and achieve strong performance after fine-tuning~\cite{wang2023foundation, wang2025improving, tian2025endomamba, dermyer2025endodino}. However, these methods remain task-specific and lack explicit multi-step evidence integration. In parallel, medical MLLMs including LLaVA-Med and ColonGPT provide flexible multimodal reasoning for visual question answering and report generation~\cite{li2023llava, ji2024frontiers}. Despite improved adaptability, current MLLMs still struggle with fine-grained visual grounding in endoscopic scenes, where subtle local cues are critical for reliable diagnosis~\cite{chang2025medheval}. Thus, existing approaches either emphasize strong local perception without coherent reasoning, or flexible reasoning without sufficiently reliable grounding.

\subsection{Medical Agents}
Medical agents have emerged as a promising paradigm for integrating reasoning with specialized diagnostic tools. General frameworks such as MMedAgent and MedOrch demonstrated the feasibility of multimodal orchestration in medical workflows~\cite{li2024mmedagent, he2025medorch}. More recent systems, including MedRAX, CARE, and MedAgent-Pro, further improve diagnostic flexibility through iterative planning, tool execution, and evidence-guided reasoning~\cite{fallahpour2025medrax, du2026care, wang2025medagent}. However, most current agents do not explicitly verify whether newly acquired evidence is sufficiently grounded before updating the reasoning trajectory. Consequently, early perception errors may propagate through later reasoning steps, which is especially problematic in endoscopic diagnosis where subtle findings strongly influence downstream decisions.

\subsection{Medical Benchmarks and Endoscopic Evaluation}
Recent medical benchmarks have increasingly shifted toward realistic clinical workflows and system-level reasoning evaluation. For instance, MedAgentBench assesses agents in simulated electronic health record environments, while AgentClinic focuses on iterative doctor-patient interaction and tool-assisted diagnostic reasoning~\cite{jiang2025medagentbench, schmidgall2024agentclinic}. In contrast, existing endoscopic benchmarks, including Kvasir-VQA, ColonINST, and EndoBench, are still largely limited to isolated perceptual subtasks or short-form question answering~\cite{gautam2024kvasir, ji2024frontiers, liu2025comprehensive}. Different from these benchmarks, EndoAgentBench evaluates the complete diagnostic workflow, spanning fine-grained lesion localization, multimodal evidence integration, and holistic report synthesis, thereby enabling unified assessment of both perceptual accuracy and clinical reasoning consistency.

\begin{figure*}[t]
    \centering
    \includegraphics[width=0.9\textwidth]{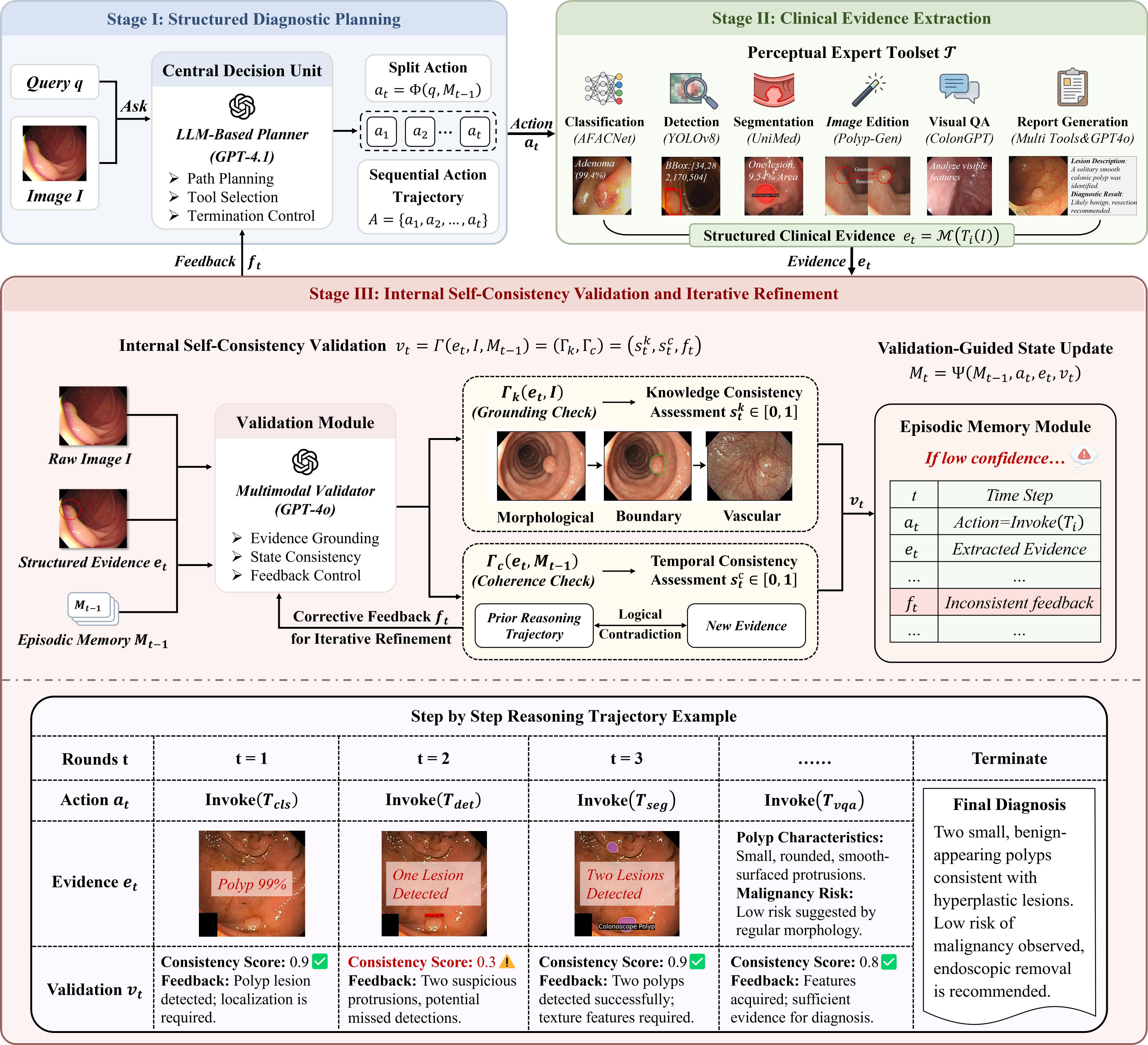}
    \caption{EndoCogniAgent framework overview. Features a closed-loop workflow in three stages: (I) planning, (II) evidence extraction, and (III) self-consistency validation. The bottom panel shows a reasoning trajectory refining the diagnostic state for final report generation.}
    \label{fig:method}
\end{figure*}

\section{Method}
EndoCogniAgent (Fig.~\ref{fig:method}) formulates endoscopic diagnosis as a closed-loop reasoning process with validation-guided state update. The central idea is to integrate image-derived observations and their textual interpretations through consistency checking, with the resulting feedback guiding how newly acquired evidence informs the evolving diagnostic state. In this formulation, the original image retains the primary visual evidence, including lesion appearance, local regions, and spatial relationships; expert-tool outputs are converted into textual evidence that records diagnostic claims, quantitative predictions, and their tool provenance; and fusion is performed by testing whether each textual claim is supported by the image and coherent with the accumulated evidence. The resulting consistency scores and corrective feedback determine how the observation is used in subsequent reasoning, rather than treating either modality as an interchangeable representation of the other.

Specifically, EndoCogniAgent performs iterative diagnosis through three sequential stages at each reasoning round, as summarized in Algorithm~\ref{alg:endocogniagent}:
(i) \textbf{Structured Diagnostic Planning}, which determines which type of clinical evidence should be acquired next based on the current diagnostic state;
(ii) \textbf{Clinical Evidence Extraction}, which invokes perceptual expert tools to obtain the corresponding quantitative or semantic evidence from the input image; and
(iii) \textbf{Internal Self-Consistency Validation}, which assesses the visual support and coherence of newly acquired evidence and provides feedback for the state update.
The updated state, including observations and validation feedback, is then fed back to guide the next planning step, yielding a closed-loop process for iterative diagnostic reasoning.

\subsection{Structured Diagnostic Planning}

Structured diagnostic planning formulates endoscopic diagnosis as a state-conditioned action-selection problem over evidence-acquisition operations. Instead of directly generating a final conclusion from the input image and query, the planner determines which type of clinical evidence to acquire next according to the evolving diagnostic state. This design converts diagnosis from one-shot prediction into an explicit sequential decision process, where each step corresponds to a deliberate evidence-acquisition action and the resulting trajectory can be explicitly traced throughout multi-step reasoning.

To realize this process, we introduce a \textbf{Central Decision Unit (CDU)} that governs the diagnostic trajectory by selecting the next action at each reasoning round. The CDU takes as input the user query $q$ and the evolving diagnostic state stored in the \textbf{Episodic Memory Module (EMM)} $M_{t-1}$, and outputs either an evidence acquisition action or a termination decision. Formally, at step $t$, the planner generates
\begin{equation}
a_t = \Phi(q, M_{t-1}),
\end{equation}
where $a_t$ is selected from the action space
\[
\mathcal{A} = \{ \text{Invoke}(T_i) \mid T_i \in \mathcal{T} \} \cup \{ \text{Terminate} \}.
\]
Here, $\text{Invoke}(T_i)$ activates a specific perceptual expert tool $T_i$ for targeted evidence extraction, while $\text{Terminate}$ indicates that the accumulated evidence is sufficient for response synthesis. In our implementation, the CDU is instantiated with GPT-4.1 as an LLM-based planner.

This formulation makes diagnostic planning explicit at the action level. Each decision is grounded in the current diagnostic state and recorded as part of the reasoning trajectory, yielding a traceable sequence of evidence-acquisition operations rather than an implicit chain of hidden intermediate steps. More importantly, this design establishes the structural interface required by subsequent stages of the framework. By representing evidence acquisition as explicit actions, newly obtained observations can be validated and selectively incorporated into the evolving diagnostic state, which is essential for suppressing hallucinated evidence and mitigating downstream error accumulation.

\begin{algorithm}[t]
\caption{Cognitive Reasoning Workflow}
\label{alg:endocogniagent}
\begin{algorithmic}[1]
    \REQUIRE User query $q$, endoscopic image $I$, perceptual expert toolset $\mathcal{T}$, max iterations $\tau_{max}$
    \ENSURE Final diagnostic response $R$
    \STATE Initialize episodic memory $M_0 \leftarrow \emptyset$

    \FOR{$t = 1$ \TO $\tau_{max}$}
        \STATE \COMMENT{\textbf{Stage I: Structured Diagnostic Planning}}
        \STATE $a_t \leftarrow \Phi(q, M_{t-1})$

        \IF{$a_t = \text{Terminate}$}
            \STATE \COMMENT{\textbf{Final Response Synthesis}}
            \STATE $R \leftarrow \text{Synthesize}(q, M_{t-1})$
            \RETURN $R$
        \ENDIF

        \STATE \COMMENT{\textbf{Stage II: Clinical Evidence Extraction}}
        \STATE Parse tool $T_i$ from action $a_t$
        \STATE $e_t \leftarrow \mathcal{M}(T_i(I))$

        \STATE \COMMENT{\textbf{Stage III: Internal Self-Consistency Validation}}
        \STATE $v_t \leftarrow \Gamma(e_t, I, M_{t-1})$

        \STATE \COMMENT{\textbf{Memory Update}}
        \STATE $M_t \leftarrow \Psi(M_{t-1}, a_t, e_t, v_t)$
    \ENDFOR

    \STATE \textbf{return} $\text{Synthesize}(q, M_{\tau_{max}})$
\end{algorithmic}
\end{algorithm}

\subsection{Clinical Evidence Extraction}

Clinical evidence extraction maps planning actions to explicit evidence observations for subsequent validation and state update. Instead of continuing to reason on the raw image, the framework invokes a task-specific perceptual expert to extract the corresponding observation at each step. This functionally decouples diagnostic planning from fine-grained visual analysis. The planner determines what evidence is needed, while the perceptual layer provides the corresponding observation through specialized models. Such decoupling is important because clinically relevant evidence in endoscopy is heterogeneous, ranging from spatial observations such as lesion location to semantic ones such as lesion category or narrative description. A single end-to-end model does not provide explicit control over which type of evidence has been acquired at each reasoning step.

To support unified reasoning across heterogeneous perceptual outputs, we introduce a \textbf{Perceptual Expert Toolset (PET)} $\mathcal{T}$. Given a planning action $a_t = \text{Invoke}(T_i)$, the selected expert tool $T_i \in \mathcal{T}$ is applied to the input image $I$, and its raw output is transformed into structured clinical evidence through a promptification function
\begin{equation}
e_t = \mathcal{M}(T_i(I)).
\end{equation}
Here, $\mathcal{M}(\cdot)$ maps heterogeneous tool-specific outputs, such as class labels, bounding boxes, segmentation masks, or free-form descriptions, into a unified structured textual evidence representation. Each evidence unit retains its source tool and the content needed for verification, such as a class probability, spatial coordinate, mask-derived measurement, or semantic description. The textual representation therefore serves as a common reasoning interface rather than a replacement for the image: it makes heterogeneous claims comparable, while the original image preserves the visual details and spatial context against which those claims are checked in the subsequent fusion stage.

The PET comprises the following clinically aligned expert tools, each responsible for extracting or generating a distinct type of diagnostic evidence.
\begin{itemize}
    \item \textbf{Classification ($T_{cls}$):} AFACNet~\cite{wang2023adaptive} provides lesion-level classification, identifying categories such as polyps, adenomas, and carcinomas.
    \item \textbf{Detection ($T_{det}$):} A fine-tuned YOLOv8~\cite{varghese2024yolov8} localizes suspicious regions and provides spatial constraints for region-level reasoning.
    \item \textbf{Segmentation ($T_{seg}$):} UniMed~\cite{wang2024universal} performs pixel-level delineation of lesions and surgical instruments, providing precise boundary information for downstream reasoning.
    \item \textbf{Visual Question Answering ($T_{vqa}$):} ColonGPT provides semantic interpretation by answering clinically relevant image-based questions.
    \item \textbf{Image Editing ($T_{edit}$):} Polyp-Gen~\cite{liu2025polyp} supports counterfactual or educational analysis through lesion synthesis and removal.
    \item \textbf{Report Generation ($T_{mrg}$):} GPT-4o consolidates accumulated multi-source evidence to support the generation of standardized diagnostic reports.
\end{itemize}

Each invoked tool contributes a candidate observation, and all observations are converted into the same evidence space before later validation. As a result, the framework does not treat perceptual outputs as final truths. Instead, it treats them as candidate evidence units whose visual grounding and cross-step coherence are assessed to guide their use in later reasoning.

\subsection{Internal Self-Consistency Validation and Iterative Refinement}

Internal self-consistency validation assesses newly acquired textual evidence against both the input image and prior observations. Rather than treating tool outputs as final truths, this stage checks their visual support and identifies conflicts that may require further examination. Validation therefore connects evidence integration with corrective feedback, guiding subsequent reasoning as the diagnostic state evolves.

\subsubsection{Validation Formulation}

The validation module provides image-support and cross-step consistency assessments for evidence integration. In our implementation, ISCV uses GPT-4o with access to the original endoscopic image. Given the newly extracted evidence $e_t$, the original endoscopic image $I$, and the current diagnostic state $M_{t-1}$, the module produces a structured validation signal:
\begin{equation}
v_t = \Gamma(e_t, I, M_{t-1}).
\end{equation}

The operator $\Gamma$ implements the explicit visual--textual fusion step. The multimodal validator jointly receives the original endoscopic image $I$, the current tool-derived textual evidence $e_t$, and the recent diagnostic state $M_{t-1}$. It returns a structured validation record $v_t = (s_t^k, s_t^c, f_t)$, consisting of a visual-grounding score, a cross-step consistency score, and corrective feedback. Because the original image is provided directly to the validator, it can examine whether the diagnostic claims encoded in $e_t$ are supported by visible morphology, lesion boundaries, vascular patterns, and spatial context. The diagnostic state provides the complementary reasoning context needed to identify conflicts with previously accumulated findings. The three components are defined as follows:

\begin{itemize}
\item \textbf{Knowledge Consistency ($s_t^k \in [0,1]$):} This assessment quantifies the extent to which $e_t$ is supported by visual evidence in the input image $I$, defined as $s_t^k = \Gamma_k(e_t, I)$. It evaluates whether semantic claims align with observable visual cues, such as morphology, vascular patterns, and boundary textures. Low scores for unsupported or contradictory assertions help limit hallucinated evidence.

\item \textbf{Temporal Consistency ($s_t^c \in [0,1]$):} This assessment quantifies the coherence between $e_t$ and the previously validated diagnostic state $M_{t-1}$, defined as $s_t^c = \Gamma_c(e_t, M_{t-1})$. It detects logical inconsistencies or unsupported hypothesis shifts that may disrupt the reasoning trajectory. Importantly, temporal consistency is used to identify conflicts that require further verification rather than to force new evidence to agree with the existing state.

\item \textbf{Corrective Feedback ($f_t$):} When inconsistencies are detected (i.e., low $s_t^k$ or $s_t^c$), $f_t$ provides natural language guidance for subsequent reasoning. Unlike scalar scores, $f_t$ serves as an interface to the planner, specifying how to resolve conflicts by re-examining visual regions or invoking alternative expert tools.
\end{itemize}

\subsubsection{Validation-Guided State Update}
The validation outcome is incorporated into memory through the state transition
\begin{equation}
M_t = \Psi(M_{t-1}, a_t, e_t, v_t).
\end{equation}
This formulation makes state update explicitly dependent on validation outcome rather than on extracted evidence alone. To operationalize this update, each newly acquired observation is assigned one of two evidence states: \textsc{Validated} or \textsc{Pending}. An observation is marked as \textsc{Validated} when it is sufficiently supported by the original image and remains coherent with the accumulated diagnostic evidence. Validated observations are incorporated into the active diagnostic state and directly condition subsequent planning. An observation is marked as \textsc{Pending} when its visual support is insufficient or when it conflicts with previously accumulated findings. Pending observations are not discarded; instead, they are retained in episodic memory together with their consistency scores, tool provenance, and corrective feedback, allowing them to be revisited as complementary evidence becomes available.

The evidence state further determines how the planner proceeds. For a \textsc{Pending} observation, the corrective feedback identifies the unsupported or conflicting component and redirects the planner toward additional verification, such as re-examining the relevant image region or invoking an alternative expert tool that provides complementary evidence. For example, an uncertain semantic description can be checked through lesion detection, segmentation-based morphology, or another task-specific observation. In this way, the state transition distinguishes evidence that can immediately support later decisions from evidence that should first undergo targeted cross-verification.

\subsubsection{Iterative Refinement and Termination}
The updated memory $M_t$ is fed back to the planner to guide the next reasoning round, yielding a closed-loop process of planning, extraction, validation, and state update. Refinement is therefore driven by the evolving reliability of the accumulated diagnostic state rather than by a fixed iteration pattern. A new observation is not rejected solely because it contradicts an earlier finding. Instead, a visually supported but temporally inconsistent observation remains \textsc{Pending} and triggers further verification. If subsequently acquired image-grounded evidence supports the new observation, the planner can revise the current diagnostic hypothesis while retaining the earlier finding and its correction history in episodic memory. This design allows reliable new evidence to correct an erroneous intermediate conclusion without sacrificing the traceability of the reasoning process.

To facilitate traceability, each reasoning round is recorded in the \textbf{Episodic Memory Module (EMM)} as a structured entry containing the action $a_t$, extracted evidence $e_t$, validation signal $v_t$, and any corrective feedback $f_t$. This design preserves a complete and auditable reasoning trace, which not only supports transparent diagnosis and controlled iterative refinement, but also provides a reliable basis for termination. The reasoning process terminates when the planner outputs $a_t = \text{Terminate}$, indicating that the currently validated state is sufficient to support response synthesis. The final diagnostic response is then generated from the terminal state $M_\tau$ rather than from any single tool output, ensuring that the conclusion is grounded in a progressively validated evidence trajectory.

\begin{figure*}[t]
    \centering
    \includegraphics[width=0.9\linewidth]{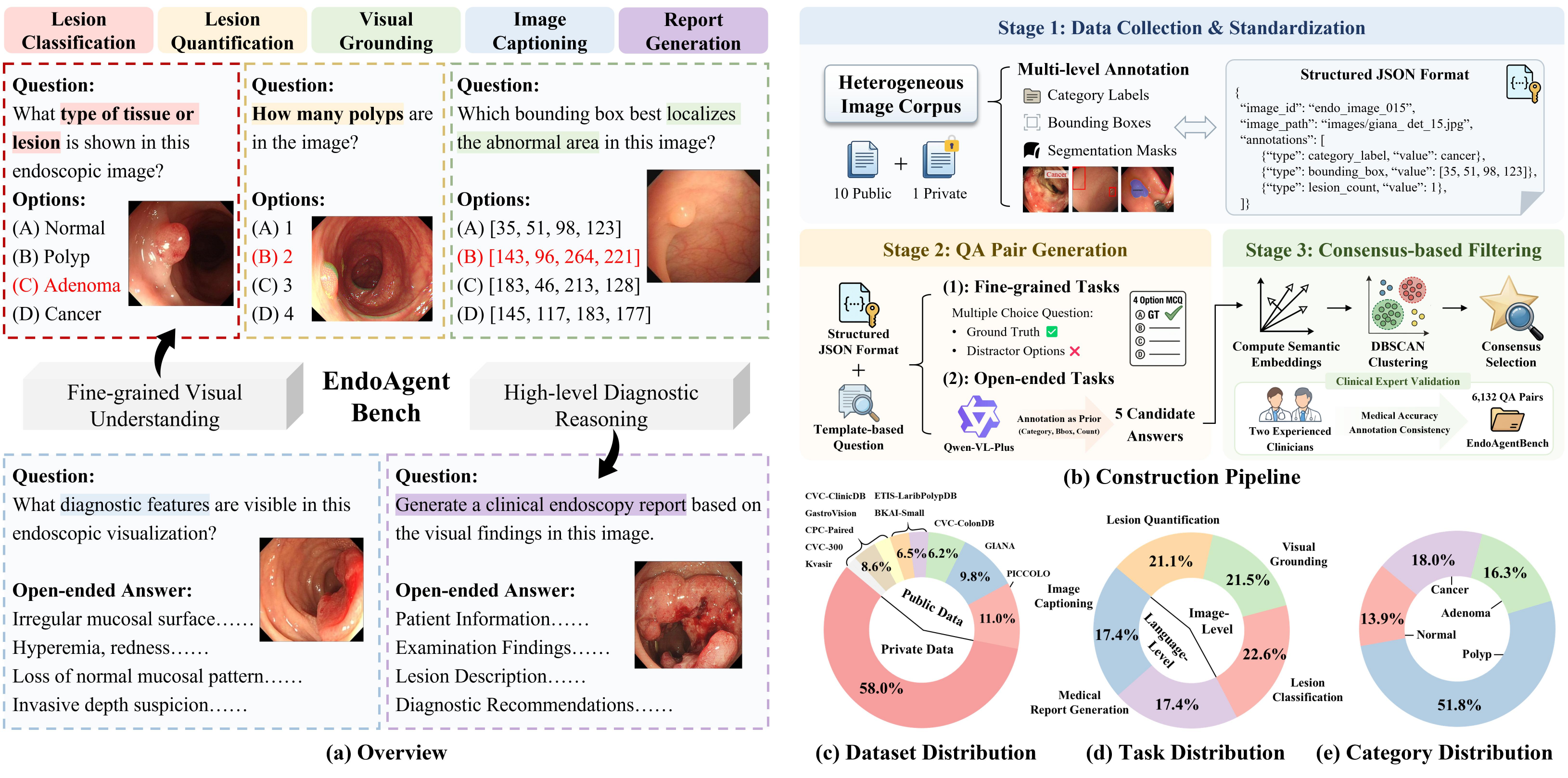}
    \caption{Overview of EndoAgentBench. (a) Core tasks spanning fine-grained perception to high-level reasoning. (b) Evidence-driven pipeline featuring automated QA generation and expert consensus. (c)-(e) Quantitative distributions across 11 datasets, task types, and categories.}
    \label{fig:Benchmark}
\end{figure*}

\section{EndoAgentBench}
EndoAgentBench (Fig.~\ref{fig:Benchmark}) is organized around two complementary capability dimensions and comprises 6,132 visual question-answer pairs from 11 endoscopic datasets. The first evaluates fine-grained visual perception, including lesion classification, lesion quantification, and visual grounding, which assess an agent's capacity to accurately recognize, count, and localize abnormal findings. The second targets high-level diagnostic reasoning through image captioning and report generation, which require synthesizing discrete visual cues and producing clinically coherent textual interpretations. Together, these tasks form a unified evaluation chain from perceptual evidence to clinical decision.

\subsection{Data Collection and Distribution}

\textbf{Dataset} (Fig.~\ref{fig:Benchmark} (b)).
The collected data includes ten widely used public endoscopic image datasets: BKAI-Small~\cite{ngoc2021neounet}, GastroVision~\cite{jha2023gastrovision}, CVC-300~\cite{bernal2012towards}, CVC-ClinicDB~\cite{bernal2015wm}, CVC-ColonDB~\cite{tajbakhsh2015automated}, Kvasir-SEG~\cite{jha2019kvasir}, ETIS-LaribPolypDB~\cite{silva2014toward}, CPC-Paired~\cite{wang2021colorectal}, PICCOLO~\cite{sanchez2020piccolo}, and GIANA~\cite{guo2019giana}, along with a private clinical dataset annotated by expert clinicians (Ethics Approval: 2025-SR-881). In total, public datasets comprise 42.0\% and private clinical data contribute 58.0\%, providing a diverse set of data sources for evaluation.

\noindent\textbf{Category} (Fig.~\ref{fig:Benchmark} (c)).
All samples are categorized into four lesion types: normal (13.9\%), polyp (51.8\%), adenoma (16.3\%), and cancer (18.0\%).
Abnormal cases collectively constitute over two-thirds of the benchmark, providing sufficient overall coverage of both common findings and clinically critical conditions required for robust endoscopic diagnostic evaluation.

\noindent\textbf{Task} (Fig.~\ref{fig:Benchmark} (a, d)).
The benchmark tasks are structured to assess two major capability dimensions: fine-grained visual perception (65.2\%) and high-level diagnostic reasoning (34.8\%). Five core subtasks are defined within these dimensions: (1) Lesion classification (22.6\%): identify the type of tissue or lesion in the image; (2) Lesion quantification (21.1\%): count the number of lesions present; (3) Visual grounding (21.5\%): localize abnormal areas via precise bounding box selection; (4) Image captioning (17.4\%): describe diagnostic features in open-ended text; (5) Report generation (17.4\%): generate comprehensive clinical endoscopy reports.

\subsection{Question-Answer Pair Generation}
To systematically construct EndoAgentBench, we design an automated pipeline for generating high-quality question-answer (QA) pairs. For fine-grained visual perception tasks, each image is associated with a diverse set of question templates, while candidate answers are automatically derived from corresponding image annotations. Specifically, for lesion classification, standard answers are generated according to category labels; for visual grounding, correct options are derived from bounding box annotations, and challenging distractor options are automatically created to increase task difficulty; for lesion quantification, stratified sampling of samples with varying numbers of lesions ensures coverage of complex scenarios.
For high-level diagnostic reasoning tasks, we adopt a self-consistency-based framework~\cite{wang2022self} to enhance the reliability of reference answers. Specifically, Qwen-VL-Plus generates $K=5$ candidate responses for each image-question pair under the same prompt, with lesion information provided to support clinically relevant generation. To determine the final reference answer, semantic embeddings are computed for all candidates, which are then grouped using a clustering strategy based on DBSCAN~\cite{ester1996density} to identify the dominant semantic consensus. The representative response from the largest cluster is selected as the reference answer. Finally, a subset of the generated QA pairs is reviewed by two experienced clinicians, with disagreements resolved through discussion to reach consensus.

\begin{table*}[t]
\centering
\caption{Performance comparison of different MLLMs and agentic frameworks on fine-grained perception and high-level diagnostic reasoning tasks. Bold and underlined values indicate the best and second-best performance, respectively.}
\label{tab:combined_performance}
\small
\begin{tabular}{l l ccc c ccc}
\toprule
\multirow{2}{*}{\textbf{Category}} & \multirow{2}{*}{\textbf{Model}}
  & \multicolumn{4}{c}{\textbf{Perception Tasks (Accuracy \%)}}
  & \multicolumn{3}{c}{\textbf{Reasoning Tasks (CAR \%)}} \\
\cmidrule(lr){3-6} \cmidrule(lr){7-9}
  & & \textbf{L. Class.} & \textbf{L. Quant.} & \textbf{V. Ground.} & \textbf{Avg.}
  & \textbf{CAP} & \textbf{MRG} & \textbf{Avg.} \\
\midrule

\multirow{3}{*}{Medical MLLMs}
& ColonGPT  & 33.50 & 43.09 & 5.23  & 27.29 & 52.91 & 20.54 & 36.71 \\
& LLaVA-Med & 30.62 & 6.49  & 27.75 & 21.87 & 3.20  & 2.25  & 2.72 \\
& HuatuoGPT & 58.36 & 75.60 & 36.85 & 56.85 & 41.07 & 20.54 & 30.80 \\
\midrule

\multirow{4}{*}{General MLLMs}
& Step-1o     & 51.80 & 59.77 & 54.06 & 55.12 & 61.18 & \underline{54.69} & 57.93 \\
& Yi-Vision   & 46.90 & 86.41 & 51.71 & 61.27 & 47.09 & 25.52 & 36.29 \\
& GPT-4o      & \underline{66.21} & 83.94 & 38.29 & 62.75 & \underline{76.03} & 50.19 & 63.10 \\
& Gemini 2.5 Pro & 63.40 & 70.12 & 76.27 & 69.82 & 53.29 & 35.83 & 44.55 \\
\midrule

\multirow{3}{*}{Agents}
& CoT   & 65.35 & 72.74 & 37.76 & 58.65 & \textbf{76.79} & 53.75 & \underline{65.26} \\
& ReAct & 61.53 & \underline{86.56} & \underline{82.11} & \underline{76.41} & 51.69 & 46.81 & 49.25 \\
& Reflexion & 42.22 & 45.48 & 75.66 & 54.30 & 58.74 & 52.35 & 55.54 \\
\midrule

\textbf{Our Method} & \textbf{EndoCogniAgent} & \textbf{75.29} & \textbf{92.90} & \textbf{88.17} & \textbf{85.23} & 64.00 & \textbf{78.24} & \textbf{71.13} \\
\bottomrule
\end{tabular}
\end{table*}

\section{Experiments}
\subsection{Evaluation Metrics}
For fine-grained visual perception tasks, such as lesion classification, quantification, and grounding, we adopt \textit{accuracy} as the evaluation metric, as these tasks require precise perception of subtle visual cues where exact correctness is critical.
For high-level diagnostic reasoning tasks, traditional automatic metrics like BLEU or ROUGE are inadequate to capture clinical validity. Following recent medical agent benchmarks~\cite{li2024mmedagent, bansal2024medmax}, we adopt an \textbf{LLM-as-a-Judge} evaluation paradigm, using GPT-4o-mini to score each model output on a 1--10 scale based on diagnostic accuracy, report completeness, and clinical coherence. Inspired by the rigorous evaluation framework of Med-PaLM~\cite{singhal2023large}, we further report the \textbf{Clinical Acceptance Rate (CAR)}, defined as the proportion of responses receiving a judge score of at least 6:
\begin{equation}
\mathrm{CAR} = \frac{1}{N}\sum_{i=1}^{N}\mathbb{I}(s_i \geq 6),
\end{equation}
where \(s_i\) denotes the judge score of the \(i\)-th sample and \(\mathbb{I}(\cdot)\) is the indicator function. CAR reflects the proportion of responses that meet the predefined acceptance threshold under the LLM-as-a-Judge protocol, rather than measuring lexical similarity to the reference answer. It therefore provides a task-level summary of clinically oriented response quality under this evaluation setting.

\subsection{Comparison with State-of-the-art Methods}
We compare EndoCogniAgent with a broad range of state-of-the-art models, including general-purpose MLLMs such as Step-1o, Yi-Vision~\cite{young2024yi}, GPT-4o~\cite{hurst2024gpt}, and Gemini 2.5 Pro~\cite{comanici2025gemini}, medical MLLMs including LLaVA-Med~\cite{li2023llava}, HuatuoGPT~\cite{chen2024huatuogpt}, and ColonGPT~\cite{ji2024frontiers}, as well as three representative agentic reasoning frameworks: Chain-of-Thought (CoT)~\cite{wei2022chain}, ReAct~\cite{yao2022react}, and Reflexion~\cite{shinn2023reflexion}. These agentic baselines also provide a functional comparison of evidence-integration strategies. ReAct performs iterative tool invocation and directly aggregates the resulting observations without the proposed consistency validation, whereas Reflexion revises the reasoning trajectory through textual self-reflection without explicitly re-grounding each reflected claim in the original image. EndoCogniAgent instead validates structured textual evidence jointly against the source image and, in the full model, the accumulated validated state. All methods are evaluated under the same protocol on EndoAgentBench.

\noindent\textbf{Performance on fine-grained visual perception tasks.}
EndoCogniAgent achieves state-of-the-art performance across all fine-grained perception subtasks. As shown in Table~\ref{tab:combined_performance}, it achieves 75.29\% lesion classification accuracy, 92.90\% lesion quantification accuracy, and 88.17\% visual grounding accuracy, resulting in an overall visual score of 85.23\%. Notably, EndoCogniAgent surpasses the strongest baseline, ReAct, by 8.82 percentage points (76.41\%), and consistently outperforms all general-purpose and medical MLLMs as well as the other agentic baselines. Because ReAct already uses iterative planning and the same type of expert-tool evidence aggregation, this comparison indicates that tool composition alone does not account for the improvement. The additional gain is consistent with the proposed use of image-grounded validation and validation-guided state updates, whose individual effects are examined in Table~\ref{tab:comprehensive_ablation}.

\noindent\textbf{Performance on high-level diagnostic reasoning tasks.}
For high-level diagnostic reasoning, EndoCogniAgent achieves the highest overall CAR of 71.13\%, outperforming CoT, the strongest baseline, by 5.87 percentage points (65.26\%). Specifically, it attains CARs of 64.00\% and 78.24\% on the Image Captioning (CAP) and Medical Report Generation (MRG) tasks, respectively. While CoT performs better on CAP (76.79\% versus 64.00\%), EndoCogniAgent substantially outperforms all baselines on MRG, exceeding Step-1o, the strongest baseline on this task, by 23.55 percentage points (54.69\%). This task-dependent pattern suggests that the proposed closed loop is particularly beneficial when long-form generation requires multiple tool-derived claims to remain visually grounded and mutually coherent, but does not uniformly improve every language task. ReAct provides a comparison with direct multi-tool aggregation, while Reflexion offers a complementary comparison with textual self-refinement without explicit image-based claim verification. Their lower overall reasoning CARs of 49.25\% and 55.54\%, respectively, further support the value of reconnecting intermediate textual claims to the source image before incorporating them into the diagnostic state.

\begin{table*}[t]
\centering
\caption{Systematic ablation of EndoCogniAgent. The results demonstrate incremental gains from the Agentic Core (PET+CDU), ISCV, and EMM. Bold and underlined values indicate the best and second-best performance, respectively.}
\label{tab:comprehensive_ablation}
\setlength{\tabcolsep}{10pt}
\begin{tabular}{ccc cccc ccc}
\toprule
\multicolumn{3}{c}{\textbf{Modules}} & \multicolumn{4}{c}{\textbf{Perception Tasks (Accuracy \%)}} & \multicolumn{3}{c}{\textbf{Reasoning Tasks (CAR \%)}} \\
\cmidrule(lr){1-3} \cmidrule(lr){4-7} \cmidrule(l){8-10}
\textbf{PET+CDU} & \textbf{ISCV} & \textbf{EMM} & \textbf{L. Class.} & \textbf{L. Quant.} & \textbf{V. Ground.} & \textbf{Avg.} & \textbf{CAP} & \textbf{MRG} & \textbf{Avg.} \\
\midrule
\multicolumn{3}{c}{\textit{Baseline (GPT-4o)}} & 66.21 & 83.94 & 38.29 & 62.75 & \textbf{76.03} & 50.19 & 63.10 \\ \midrule

\checkmark & & & 61.53 & 86.56 & \underline{82.11} & 76.41 & 51.69 & 46.81 & 49.25 \\

\checkmark & \checkmark & & \underline{69.09} & \underline{88.57} & 76.72 & \underline{77.91} & 59.77 & \underline{70.83} & \underline{65.31} \\

\checkmark & \checkmark & \checkmark & \textbf{75.29} & \textbf{92.90} & \textbf{88.17} & \textbf{85.23} & \underline{64.00} & \textbf{78.24} & \textbf{71.13} \\
\bottomrule
\end{tabular}
\end{table*}

\subsection{Ablation Analysis}
Table~\ref{tab:comprehensive_ablation} compares four configurations with progressively added components. The GPT-4o baseline performs single-pass image-to-text reasoning. PET+CDU, identical to the ReAct baseline in Table~\ref{tab:combined_performance}, serves as the agentic-core reference, introducing expert-tool orchestration without an explicit validation gate. Adding ISCV enables image-grounded validation of structured textual evidence, while EMM further incorporates the accumulated cross-step state. In the full model, GPT-4.1 serves as the CDU planner and GPT-4o as the multimodal validator within ISCV.

\subsubsection{Effect of the Agentic Core (PET + CDU)}
Integrating the Agentic Core, which without ISCV or EMM corresponds to the standard ReAct paradigm of iterative planning and tool invocation, increases visual grounding accuracy from 38.29\% to 82.11\%, demonstrating the effectiveness of structured task decomposition and expert tool utilization. However, reasoning performance declines in the absence of intermediate evidence validation, as iterative tool invocation may introduce inconsistencies and propagate errors across reasoning steps.

\subsubsection{Effect of Internal Self-Consistency Validation (ISCV)}
Adding ISCV changes the evidence-integration mechanism from direct tool-output aggregation to image-grounded validation of structured textual claims. Relative to PET+CDU, ISCV increases average perception accuracy from 76.41\% to 77.91\% and average reasoning CAR from 49.25\% to 65.31\%; the largest gain occurs on MRG, where CAR rises from 46.81\% to 70.83\%. The improvement is not uniform across every subtask---visual grounding decreases from 82.11\% to 76.72\%---but the overall reasoning gains show that directly rechecking intermediate text against the source image is more effective than aggregating tool outputs without such grounding, especially for long-form evidence synthesis.

\subsubsection{Effect of Episodic Memory Module (EMM)}
Finally, integrating the EMM extends the fusion scope from the current image--text pair to the sequence of previously validated observations. Compared with ISCV without EMM, the full model improves average perception accuracy from 77.91\% to 85.23\% and average reasoning CAR from 65.31\% to 71.13\%. This gain suggests that retaining earlier evidence and validation feedback helps the planner maintain coherent reasoning across steps, particularly in complex diagnostic scenarios.

\begin{table*}[t]
\centering
\caption{Paired statistical comparison between EndoCogniAgent and baseline methods on the 2,130 reasoning samples. Mean Judge Score is reported as Mean $\pm$ SD on a 1--10 scale. CAR denotes the Clinical Acceptance Rate. $\Delta$ denotes the difference between EndoCogniAgent and each baseline, computed as $\Delta=\mathrm{Ours}-\mathrm{Baseline}$. All statistical tests are performed at the sample level.}
\label{tab:stat_significance}
\small
\setlength{\tabcolsep}{6pt}
\begin{tabular}{l cc cc cc}
\toprule
\multirow{2}{*}{\textbf{Model}} &
\multicolumn{2}{c}{\textbf{Performance}} &
\multicolumn{2}{c}{\textbf{Difference from Ours}} &
\multicolumn{2}{c}{\textbf{Paired Test ($p$)}} \\
\cmidrule(lr){2-3}
\cmidrule(lr){4-5}
\cmidrule(lr){6-7}
& \textbf{Mean Score} & \textbf{CAR (\%)}
& \textbf{$\Delta$ Score} & \textbf{$\Delta$ CAR}
& \textbf{Wilcoxon} & \textbf{McNemar} \\
\midrule
\textbf{EndoCogniAgent (Ours)}
& \textbf{$6.08 \pm 1.83$} & \textbf{71.13}
& -- & --
& -- & -- \\
\midrule
CoT
& $5.91 \pm 2.09$ & 65.26
& +0.172 & +5.87
& $1.2\times10^{-3}$ & $3.0\times10^{-5}$ \\

GPT-4o
& $5.73 \pm 2.04$ & 63.10
& +0.346 & +8.03
& $<10^{-3}$ & $<10^{-3}$ \\

Step-1o
& $5.64 \pm 3.31$ & 57.93
& +0.439 & +13.20
& $<10^{-3}$ & $<10^{-3}$ \\

Reflexion
& $5.39 \pm 2.52$ & 55.54
& +0.688 & +15.59
& $<10^{-3}$ & $<10^{-3}$ \\

ReAct
& $4.86 \pm 2.13$ & 49.25
& +1.217 & +21.88
& $<10^{-3}$ & $<10^{-3}$ \\

Gemini 2.5 Pro
& $5.13 \pm 2.72$ & 44.55
& +0.952 & +26.58
& $<10^{-3}$ & $<10^{-3}$ \\

ColonGPT
& $4.43 \pm 2.12$ & 36.71
& +1.652 & +34.42
& $<10^{-3}$ & $<10^{-3}$ \\

Yi-Vision
& $4.65 \pm 2.00$ & 36.29
& +1.433 & +34.84
& $<10^{-3}$ & $<10^{-3}$ \\

HuatuoGPT
& $4.10 \pm 2.01$ & 30.80
& +1.977 & +40.33
& $<10^{-3}$ & $<10^{-3}$ \\

LLaVA-Med
& $2.49 \pm 1.43$ & 2.72
& +3.595 & +68.41
& $<10^{-3}$ & $<10^{-3}$ \\
\bottomrule
\end{tabular}
\end{table*}

\subsection{Statistical Significance and Reliability Analysis}
To assess the statistical significance and reliability of the reasoning performance, we conduct paired statistical tests and estimate 95\% bootstrap confidence intervals for CAR over the 2,130 samples from the CAP and MRG tasks.

\noindent\textbf{Paired significance testing.}
We evaluate two complementary metrics on the 2,130 reasoning samples: the mean judge score (on a 1--10 scale) and the binary CAR (the proportion of scores $\geq 6$). EndoCogniAgent attains a mean judge score of $6.08\pm1.83$, which exceeds that of every baseline (Table~\ref{tab:stat_significance}). To assess whether these differences are statistically significant on a per-sample basis, we apply the paired Wilcoxon signed-rank test to the judge scores of EndoCogniAgent and each baseline. All comparisons are statistically significant ($p<0.05$), with 9 of 10 comparisons reaching $p<0.001$. Since CAR is derived from binary acceptance decisions ($s_i \geq 6$), we further apply McNemar's test to the paired accepted/rejected decisions. The results confirm that the CAR improvement of EndoCogniAgent over every baseline is statistically significant (all $p<0.001$).

\begin{figure}[t]
    \centering
    \includegraphics[width=0.9\linewidth]{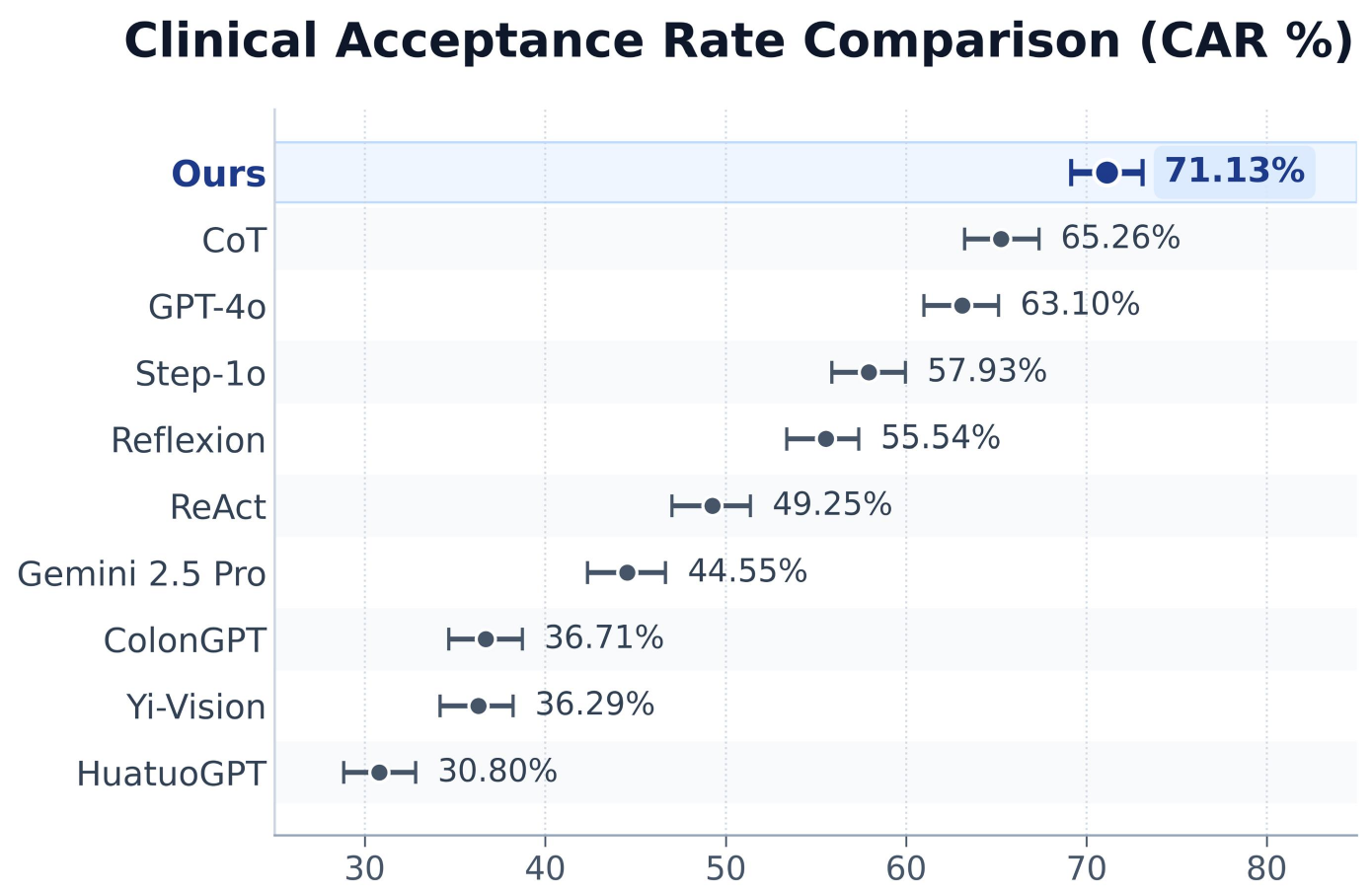}
    \caption{CAR with 95\% bootstrap confidence intervals. Points denote the observed CAR, and horizontal bars indicate the 95\% percentile bootstrap confidence intervals. LLaVA-Med (CAR 2.72\%) is omitted for visual clarity.}
    \label{fig:car_ci}
\end{figure}

\noindent\textbf{Bootstrap confidence intervals.}
To quantify the uncertainty of the CAR estimate, we compute 95\% percentile bootstrap confidence intervals by resampling the 2,130 reasoning samples with replacement $B=1{,}000$ times and taking the 2.5-th and 97.5-th percentiles of the resulting bootstrap distribution:
\begin{equation}
\mathrm{CI}_{95\%} =
\left[
Q_{0.025}\bigl(\{\mathrm{CAR}^{(b)}\}_{b=1}^{B}\bigr),
Q_{0.975}\bigl(\{\mathrm{CAR}^{(b)}\}_{b=1}^{B}\bigr)
\right].
\end{equation}
As shown in Fig.~\ref{fig:car_ci}, EndoCogniAgent achieves a CAR of 71.13\% with a 95\% bootstrap confidence interval of $[69.15\%, 73.10\%]$. The relatively narrow interval indicates low sampling uncertainty in the estimated CAR.

\begin{table*}[t]
\centering
\caption{Comparison of three-clinician consensus and automatic LLM-judge evaluation on 100 stratified-random reasoning samples. $|\Delta|$ denotes the absolute difference between the two evaluation protocols.}
\label{tab:human_eval}
\small
\setlength{\tabcolsep}{4pt}
\begin{tabular}{l ccc ccc}
\toprule
\multirow{2}{*}{\textbf{Model}} &
\multicolumn{3}{c}{\textbf{Mean score}} &
\multicolumn{3}{c}{\textbf{CAR (\%)}} \\
\cmidrule(lr){2-4} \cmidrule(lr){5-7}
& \textbf{Clinicians} & \textbf{LLM judge} & \textbf{$|\Delta|$}
& \textbf{Clinicians} & \textbf{LLM judge} & \textbf{$|\Delta|$} \\
\midrule
EndoCogniAgent (Ours) & \textbf{6.34} & \textbf{6.10} & \textbf{0.24} & \textbf{71.0} & \textbf{73.0} & \textbf{2.0} \\
GPT-4o                & 5.50 & 5.52 & 0.02 & 50.0 & 60.0 & 10.0 \\
ColonGPT              & 4.42 & 4.61 & 0.19 & 32.0 & 42.0 & 10.0 \\
\midrule
\multicolumn{7}{c}{\textbf{Overall human--LLM agreement:} Pearson's $r=0.608$; Binary agreement $=70.7\%$} \\
\bottomrule
\end{tabular}
\end{table*}

\begin{table}[t]
\centering
\caption{Paired significance and clinician-rating reliability. Wilcoxon tests assess paired score differences; senior--junior agreement averages the two comparisons with the senior clinician as reference, and $\mathrm{ICC}(2,3)$ assesses the reliability of the three-clinician mean rating.}
\label{tab:human_eval_stats}
\small
\setlength{\tabcolsep}{3pt}
\begin{tabular}{l c c}
\toprule
\multicolumn{3}{c}{\textbf{Paired significance (EndoCogniAgent vs.\ baseline)}} \\
\midrule
\textbf{Comparison} & \textbf{$\Delta$Score} & \textbf{Wilcoxon $p$} \\
vs.\ ColonGPT & $+1.92$ & $<10^{-10}$ \\
vs.\ GPT-4o   & $+0.84$ & $0.002$ \\
\midrule
\multicolumn{3}{c}{\textbf{Senior--junior agreement and consensus reliability}} \\
\midrule
\textbf{Rating target} & \textbf{Agreement metric} & \textbf{Value} \\
\multirow{2}{*}{CAR (score $\geq6$)}
& Senior--junior Agreement & $0.728$ \\
& Senior--junior Cohen's $\kappa$ & $0.442$ \\
Mean score (1--10)
& $\mathrm{ICC}(2,3)$ & $0.732$ \\
\bottomrule
\end{tabular}
\end{table}

\subsection{Human Evaluation}

To independently assess clinical response quality and validate the LLM-as-a-Judge protocol, we conducted a blinded human evaluation following established clinician-based evaluation practice for medical language models~\cite{singhal2023large}.

\noindent\textbf{Experimental design.}
Using a fixed random seed, we selected 100 reasoning items by stratified random sampling from the CAP and MRG test sets (50 per task), with representation from four diagnostic categories: normal, polyp, adenoma, and colorectal cancer. For each item, the responses generated by EndoCogniAgent, GPT-4o, and ColonGPT were randomly ordered and presented as anonymous candidates A, B, and C.
Three endoscopists, one chief physician serving as the senior clinician and two junior physicians, independently rated every response on a $1$--$10$ clinical acceptability scale while blinded to model identity and automatic judge scores. Ratings of $1$--$3$, $4$--$5$, $6$--$8$, and $9$--$10$ denoted unacceptable, borderline, clinically acceptable, and excellent responses, respectively. This protocol yielded 900 ratings for 300 model responses. For each response, the three clinician ratings were averaged to form the consensus score; a consensus score $\geq6$ was classified as clinically acceptable for calculating CAR. The senior clinician was the reference for the two senior--junior agreement comparisons. Reliability of the three-clinician mean rating was quantified using a two-way random-effects, absolute-agreement $\mathrm{ICC}(2,3)$, and paired item-level consensus scores were compared using Wilcoxon signed-rank tests.

\noindent\textbf{Human evaluation results.}
As shown in Table~\ref{tab:human_eval}, EndoCogniAgent received the highest clinician consensus mean score (6.34) and CAR (71.0\%), followed by GPT-4o (5.50; 50.0\%) and ColonGPT (4.42; 32.0\%). Thus, EndoCogniAgent improved CAR by 21.0 percentage points over GPT-4o and 39.0 percentage points over ColonGPT. The paired analysis in Table~\ref{tab:human_eval_stats} confirmed higher consensus scores for EndoCogniAgent than for GPT-4o ($\Delta\mathrm{Score}=+0.84$, $p=0.002$) and ColonGPT ($\Delta\mathrm{Score}=+1.92$, $p<10^{-10}$).

\noindent\textbf{Reliability of the clinician consensus.}
For binary clinical acceptability, the two senior--junior comparisons yielded a mean agreement of 0.728, showing that the clinicians reached the same acceptability decision for most responses. The corresponding mean Cohen's $\kappa$ was 0.442, indicating moderate agreement after accounting for agreement expected by chance. For the continuous $1$--$10$ scores, the three-clinician mean rating achieved $\mathrm{ICC}(2,3)=0.732$. Together, these results support the reliability of both the threshold-based CAR assessment and the averaged consensus score used for model comparison.

\noindent\textbf{Agreement between automatic and human evaluation.}
The LLM judge reproduced the clinician-derived model ranking. Its model-level mean scores differed from the clinician consensus by only 0.24, 0.02, and 0.19 points for EndoCogniAgent, GPT-4o, and ColonGPT, respectively; the corresponding CAR differences were 2.0, 10.0, and 10.0 percentage points. Across all 300 responses, human and LLM-judge scores were positively correlated (Pearson's $r=0.608$), and their binary acceptability decisions agreed in 70.7\% of cases. Together, these results show that the automatic judge captured the clinician-assessed performance ordering and achieved moderate response-level concordance, supporting its use for scalable evaluation alongside clinician review in practice.

\begin{figure*}[t]
    \centering
    \includegraphics[width=0.9\linewidth]{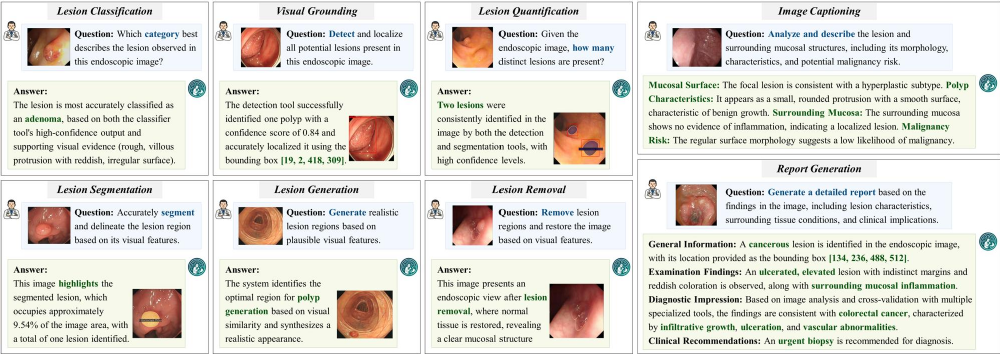}
    \caption{Multi-task performance of EndoCogniAgent across fine-grained perception and high-level clinical reasoning. Specialized expert tools enable grounded visual outputs and clinically coherent diagnostic responses across diverse scenarios.}
    \label{fig:Multi-Task}
\end{figure*}

\subsection{Multi-Task Coverage}
Fig.~\ref{fig:Multi-Task} showcases the broad task coverage of EndoCogniAgent, spanning fine-grained visual perception and high-level clinical reasoning. By coordinating specialized expert tools, the agent produces grounded visual outputs, such as segmentation masks and bounding boxes, together with coherent clinical narratives. These capabilities are integrated within a unified closed-loop architecture, enabling EndoCogniAgent to address diverse diagnostic tasks through coordinated evidence acquisition and reasoning.

\begin{figure}[t]
    \centering
    \includegraphics[width=0.9\linewidth]{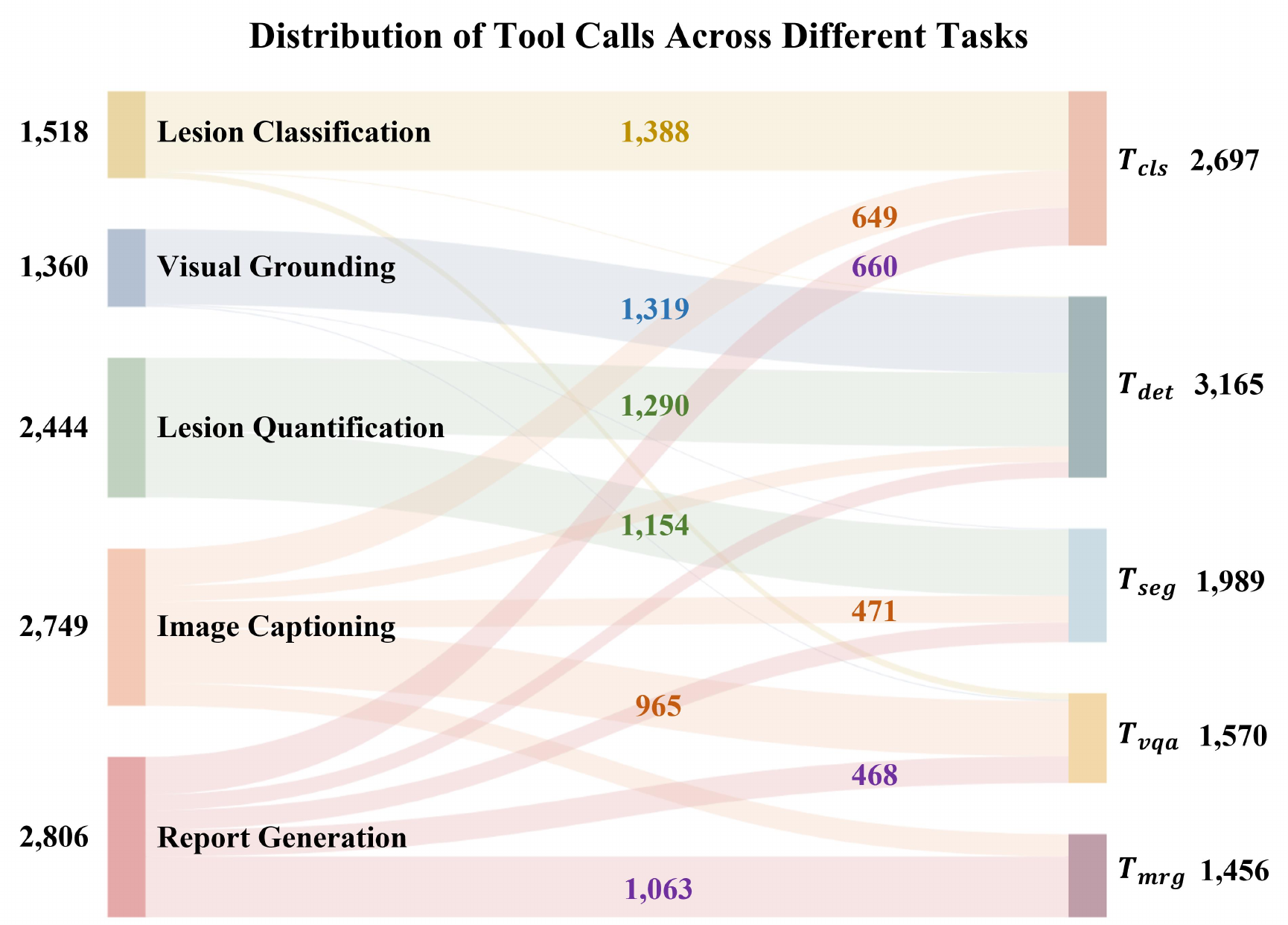}
    \caption{Sankey diagram of tool distributions. It illustrates the CDU's adaptive orchestration across tasks, with link widths proportional to tool invocation frequency.}
    \label{fig:Sankey}
\end{figure}

\subsection{Tool Orchestration Analysis}
As visualized in Fig.~\ref{fig:Sankey}, the CDU exhibits hierarchical orchestration patterns that scale with task complexity. Simple perceptual tasks show strong convergence on specialized tools: $T_{cls}$ accounts for 91.4\% of the tool calls for Lesion Classification (1,388 of 1,518), while $T_{det}$ accounts for 1,319 of the 1,360 calls for Visual Grounding. In contrast, compositional tasks such as Lesion Quantification shift toward a dual-expert strategy, balancing 1,154 detections via $T_{det}$ with 1,290 segmentations through $T_{seg}$ to achieve precise "locate-then-delineate" reasoning. This transition from single-tool convergence to multi-expert coordination is most pronounced in Report Generation, which demands peak reasoning intensity with 2,806 total calls. Here, the CDU orchestrates a complex collective of $T_{mrg}$, $T_{vqa}$, and $T_{seg}$, which contribute 1,063, 660, and 471 calls respectively. These task-dependent orchestration patterns suggest that the CDU allocates more diverse evidence-acquisition operations to complex reasoning tasks while relying more strongly on specialized tools.

\begin{table*}[t]
\centering
\caption{Performance of EndoCogniAgent with different CDU backbone MLLMs. Bold and underlined values indicate the best and second-best performance, respectively.}
\label{tab:scalability_mllm}
\small
\begin{tabular}{l cccc ccc}
\toprule
\multirow{2}{*}{\textbf{MLLM}} &
\multicolumn{4}{c}{\textbf{Perception Tasks (Accuracy \%)}} &
\multicolumn{3}{c}{\textbf{Reasoning Tasks (CAR \%)}} \\
\cmidrule(lr){2-5} \cmidrule(lr){6-8}
& \textbf{L. Class.} & \textbf{L. Quant.} & \textbf{V. Ground.} & \textbf{Avg.}
& \textbf{CAP} & \textbf{MRG} & \textbf{Avg.} \\
\midrule
GPT-4.1         & \textbf{70.63} & \textbf{95.00} & 88.33 & \textbf{83.25} & \underline{72.50} & 82.50 & 77.50 \\
Claude-Sonnet-5 & 50.63 & 92.50 & \textbf{97.50} & 77.25 & \textbf{91.25} & \textbf{88.75} & \textbf{90.00} \\
Grok-4          & \underline{69.38} & 91.67 & \underline{92.50} & \underline{83.00} & 71.25 & \underline{85.00} & \underline{78.13} \\
Gemini-2.5-Pro  & 58.75 & \underline{94.17} & 87.50 & 78.00 & 65.00 & 80.00 & 72.50 \\
\midrule
\textbf{Performance Range} & 20.00 & 3.33 & 10.00 & 6.00 & 26.25 & 8.75 & 17.50 \\
\bottomrule
\end{tabular}
\end{table*}

\begin{figure}[t]
    \centering
    \includegraphics[width=0.9\linewidth]{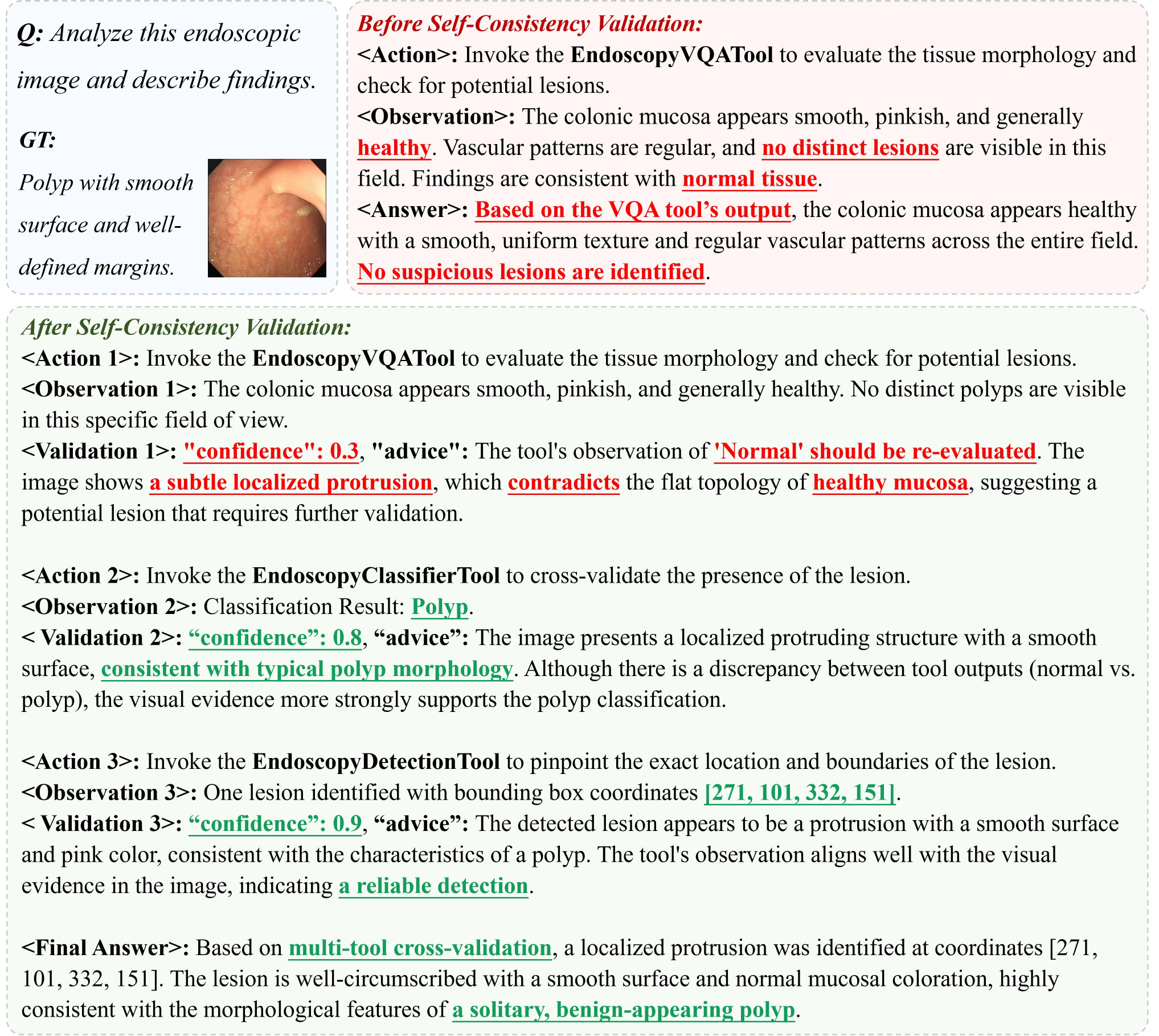}
    \caption{Iterative error correction via ISCV. EndoCogniAgent detects semantic conflicts and refines the diagnosis through progressive, validation-guided verification.}
    \label{fig:Multi-Round}
\end{figure}

\subsection{Iterative Error Correction}
The case in Fig.~\ref{fig:Multi-Round} illustrates the role of ISCV in preventing early reasoning errors from propagating. In the single-pass baseline, an erroneous VQA response describing ``normal mucosa'' is accepted without verification, leading to a missed polyp. EndoCogniAgent instead compares the evidence against the visual observation and identifies a knowledge consistency violation caused by the focal mucosal protrusion. ISCV records the disputed evidence with corrective feedback that redirects the planner toward verification. Over three iterative rounds, the agent progressively cross-validates the lesion and performs spatial localization, increasing diagnostic confidence from 0.3 to 0.9. The example illustrates how validation-guided state updates can prompt the reconsideration and correction of an unreliable intermediate observation during subsequent reasoning.

\begin{figure}[t]
    \centering
    \includegraphics[width=0.9\columnwidth]{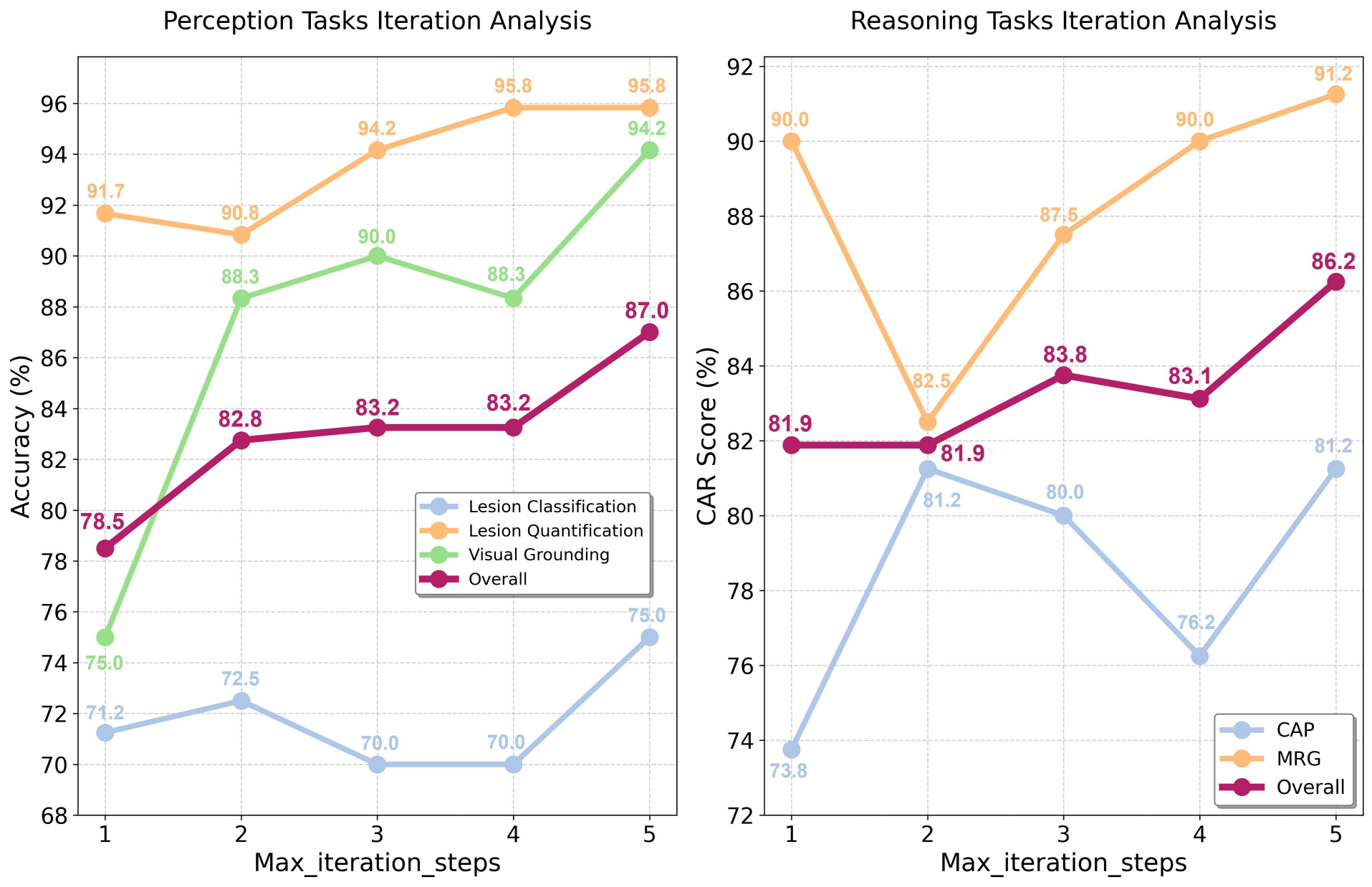}
    \caption{Impact of reasoning rounds on diagnostic performance in both perception (left) and reasoning (right) tasks.}
    \label{fig:round}
\end{figure}

\subsection{Effect of the Maximum Iteration Steps}
Fig.~\ref{fig:round} shows the effect of the maximum iteration budget on diagnostic performance. Increasing the maximum number of reasoning steps from 1 to 5 raises overall perception accuracy from 78.50\% to 87.00\% and reasoning CAR from 81.88\% to 86.25\%. Performance varies at intermediate iteration budgets, particularly on the reasoning tasks, while the five-step setting achieves the highest overall performance among the evaluated configurations. Overall performance improves as more reasoning rounds are allowed, although intermediate steps show some fluctuations.

\subsection{Generalization across Backbone MLLMs}
To assess whether the advantages of EndoCogniAgent generalize beyond a single proprietary MLLM, we substitute the CDU with four MLLMs from different model families, namely GPT-4.1, Claude-Sonnet-5, Grok-4, and Gemini-2.5-Pro, while keeping the expert toolset and self-consistency validation unchanged. Using the same uniformly sampled EndoAgentBench subset as in the iteration analysis, we evaluate the framework with different backbone MLLMs. As shown in Table~\ref{tab:scalability_mllm}, EndoCogniAgent maintains strong performance on both perception and reasoning tasks across all evaluated backbones, with average perception accuracy confined to a relatively compact 77.25\%--83.25\% interval and reasoning CAR above 72\% for every backbone. The narrowest variation occurs on tasks driven by dedicated perceptual experts, where lesion quantification and visual grounding vary by only 3.33 and 10.00 percentage points, respectively, suggesting that these tool-driven tasks are relatively less sensitive to planner choice. These results support the generalizability of EndoCogniAgent across different MLLM backbones and its extensibility to alternative planners.

\section{Conclusion}
This work introduces EndoCogniAgent, a closed-loop framework that formulates endoscopic diagnosis as a controlled state update process with self-consistency validation. The framework assigns distinct roles to the two modalities: the endoscopic image supplies local appearance and spatial evidence, while tool-derived text expresses diagnostic claims in a traceable and comparable form. ISCV fuses them by assessing whether each textual claim is supported by the image and consistent with prior findings, while EMM preserves the validated evidence and feedback for subsequent reasoning. The comparisons with ReAct and Reflexion, together with the staged ablation, show that direct tool aggregation and text-only reflection are less effective than explicit image-grounded validation for overall diagnostic reasoning, with the clearest benefit observed in report generation. We further introduce EndoAgentBench to support workflow-level evaluation across fine-grained visual perception and high-level diagnostic reasoning. Several limitations remain. The current framework relies on API-based large language models, which introduces latency for real-time deployment. The benchmark is constructed on still images rather than video sequences, and prospective clinical evaluation has not yet been conducted. Future work will also extend the framework to temporal video reasoning and validate its utility in real clinical workflows.

\printcredits

\section*{Declaration of competing interest}
The authors declare that they have no known competing financial interests or personal relationships that could have appeared to influence the work reported in this paper.

\section*{Data availability}
The public datasets used in this study are openly available from the sources cited in Appendix~A. The private clinical dataset will be made available upon publication. The code is available at \url{https://github.com/Tyyds-ai/EndoCogniAgent}.

\section*{Acknowledgements}
This work was supported in part by the National Key Research and Development Program of China (2024YFF1206700), in part by the National Natural Science Foundation of China (62371121), and in part by Jiangsu Provincial Key Manufacturing Power Province Construction Special Fund Project-``1650'' Industrial System Collaborative Research of China.

\appendix
\renewcommand{\thesection}{Appendix~\Alph{section}}
\renewcommand{\thesubsection}{\Alph{section}.\arabic{subsection}}
\section{Dataset Details}
This section provides a comprehensive overview of the EndoAgentBench dataset, which forms the foundation for evaluating multimodal models in endoscopic image analysis. We detail the data sources and their composition, describe the annotation protocols for five core diagnostic tasks, and explain the preprocessing workflows tailored for both visual and language-based tasks. 
\subsection{Data Sources and Collection}
\noindent \textbf{Data Composition}
The EndoAgentBench benchmark is constructed from a diverse set of sources, including both public and private clinical data. Table \ref{tab:data} summarizes the number of samples contributed by each data source. The majority of samples originate from a private clinical dataset, supplemented by several widely used public datasets such as BKAI-Small~\cite{ngoc2021neounet}, CPC-Paired~\cite{wang2021colorectal}, GastroVision~\cite{jha2023gastrovision}, CVC-300~\cite{bernal2012towards}, CVC-ClinicDB~\cite{bernal2015wm}, CVC-ColonDB~\cite{tajbakhsh2015automated}, Kvasir-SEG~\cite{jha2019kvasir}, ETIS-LaribPolypDB~\cite{silva2014toward}, PICCOLO~\cite{sanchez2020piccolo}, and GIANA~\cite{guo2019giana}.

\noindent \textbf{Category Distribution}
The dataset covers a wide range of lesion categories, as shown in Table~\ref{tab:cat}. The most prevalent category is polyp, followed by cancer, adenoma, and normal tissue. These categories span a clinically meaningful spectrum: normal tissue serves as a negative reference, while polyps and adenomas represent benign and precancerous lesions respectively, and cancer corresponds to malignancy.

\noindent \textbf{Data Independence and Leakage Prevention}
To ensure an unbiased evaluation, the data used to train or fine-tune the perceptual expert tools were kept strictly separate from the samples included in EndoAgentBench. No benchmark image, annotation, or question--answer pair was used for tool training, hyperparameter tuning, prompt development, model selection, or early stopping. We additionally checked the provenance of the public and private data sources when constructing the benchmark to exclude overlap with the development data of the expert tools.
For data originating from videos or repeated examinations, all frames associated with the same patient or source video were treated as a single group during data partitioning. Consequently, frames from the same patient, examination, or video were not distributed across model-development and evaluation subsets. The private clinical data were partitioned at the patient level before QA generation, and only the held-out evaluation portion was included in EndoAgentBench. These measures prevent both direct sample leakage and indirect leakage through highly correlated adjacent frames.

\subsection{Data Annotation}

For each sample in EndoAgentBench, task-specific labels were generated according to the requirements of five core diagnostic tasks:

\begin{itemize}
    \item \textbf{Lesion Classification:}  
    Each image was annotated with its lesion category (normal, polyp, adenoma, or cancer) by clinicians. For public datasets, we adopted the labels; for private data, annotation was performed by board-certified physicians (see Figure~\ref{fig:cls}).

\begin{table}[h]
\centering
\caption{Data source statistics for the entire dataset.}
\label{tab:data}
\small
\begin{tabular}{lc}
\toprule
\textbf{Data Source} & \textbf{Number of Samples} \\
\midrule
Private              & 3558 \\
BKAI-Small           & 200  \\
CPC-Paired           & 165  \\
GastroVision         & 139  \\
CVC-ColonDB          & 380  \\
ETIS-LaribPolypDB    & 196  \\
Kvasir-SEG           & 99   \\
CVC-300              & 60   \\
CVC-ClinicDB         & 62   \\
PICCOLO              & 673  \\
GIANA                & 600  \\
\midrule
\textbf{Total}       & 6132 \\
\bottomrule
\end{tabular}
\end{table}

\noindent \textbf{Task Distribution} 
In addition, the dataset supports multiple task types relevant to endoscopic diagnosis, including image captioning, report generation, lesion classification, visual grounding, and lesion quantification. The distribution of samples across these tasks is presented in Table \ref{tab:task}.

\begin{table}[h]
\centering
\caption{Task type distribution in the dataset.}
\label{tab:task}
\small
\begin{tabular}{lc}
\toprule
\textbf{Task Type}           & \textbf{Number of Samples} \\
\midrule
Lesion Classification       & 1388 \\
Visual Grounding            & 1319 \\
Lesion Quantification       & 1295 \\
Image Captioning               & 1064 \\
Report Generation           & 1066 \\
\midrule
\textbf{Total}              & 6132 \\
\bottomrule
\end{tabular}
\end{table}

\begin{table}[h]
\centering
\caption{Category distribution in the dataset.}
\label{tab:cat}
\small
\begin{tabular}{lc}
\toprule
\textbf{Category} & \textbf{Number of Samples} \\
\midrule
Normal            & 855  \\
Polyp             & 3176 \\
Adenoma           & 998  \\
Cancer            & 1103 \\
\midrule
\textbf{Total}    & 6132 \\
\bottomrule
\end{tabular}
\end{table}

\begin{figure}[h]
    \centering
    \includegraphics[width=0.8\linewidth]{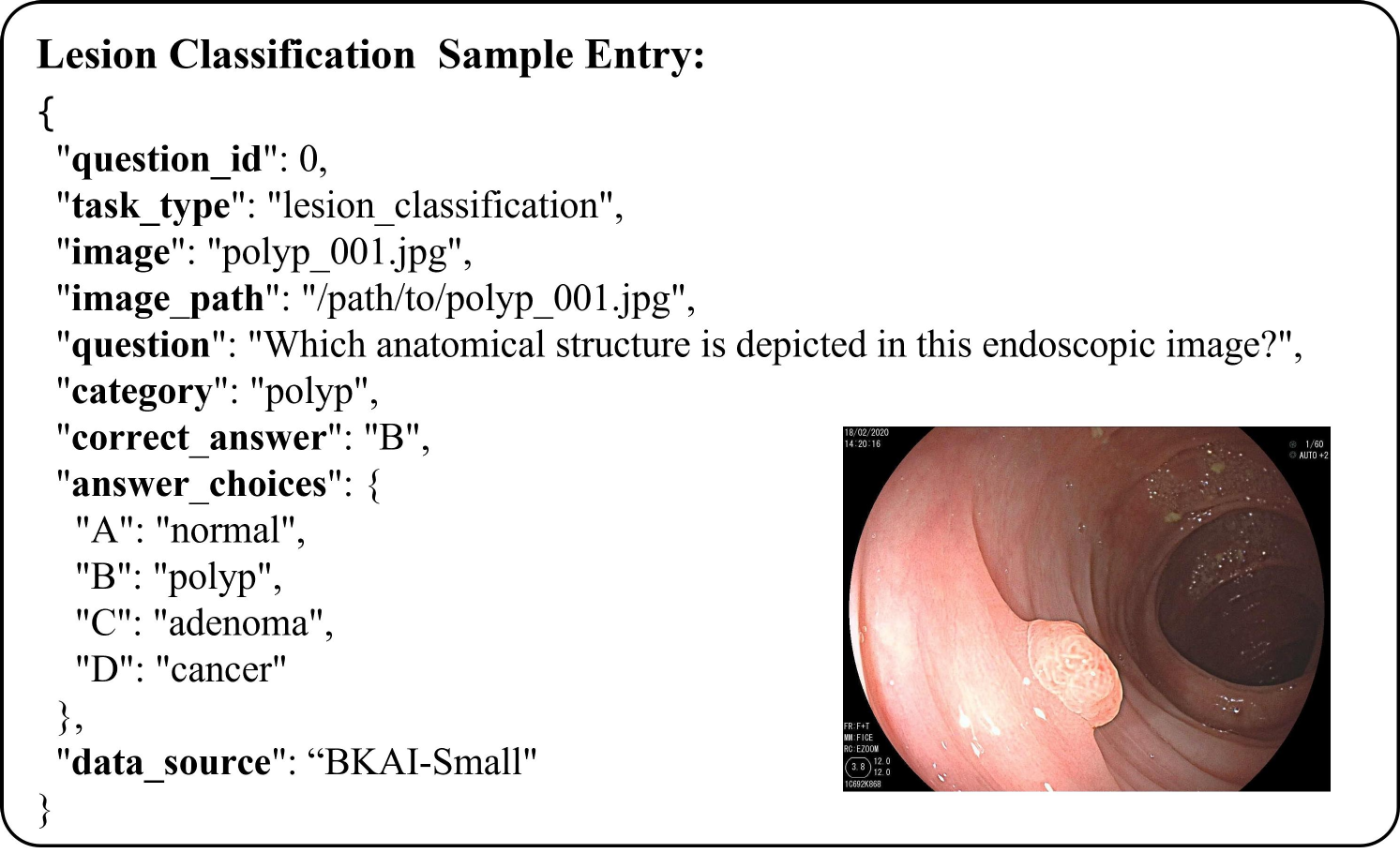}
    \caption{Lesion Classification annotation example.}
    \label{fig:cls}
\end{figure}
    
    \item \textbf{Visual Grounding:}  
    Each image containing lesions was annotated with bounding box coordinates to localize the lesion area. For public datasets, we adopted the official bounding box annotations when available; if only segmentation masks were provided, the minimum enclosing rectangles of connected components were extracted using standard image processing techniques. For private datasets, all bounding boxes were annotated by expert clinicians (see Figure~\ref{fig:vg}).

\begin{figure}[h]
    \centering
    \includegraphics[width=0.8\linewidth]{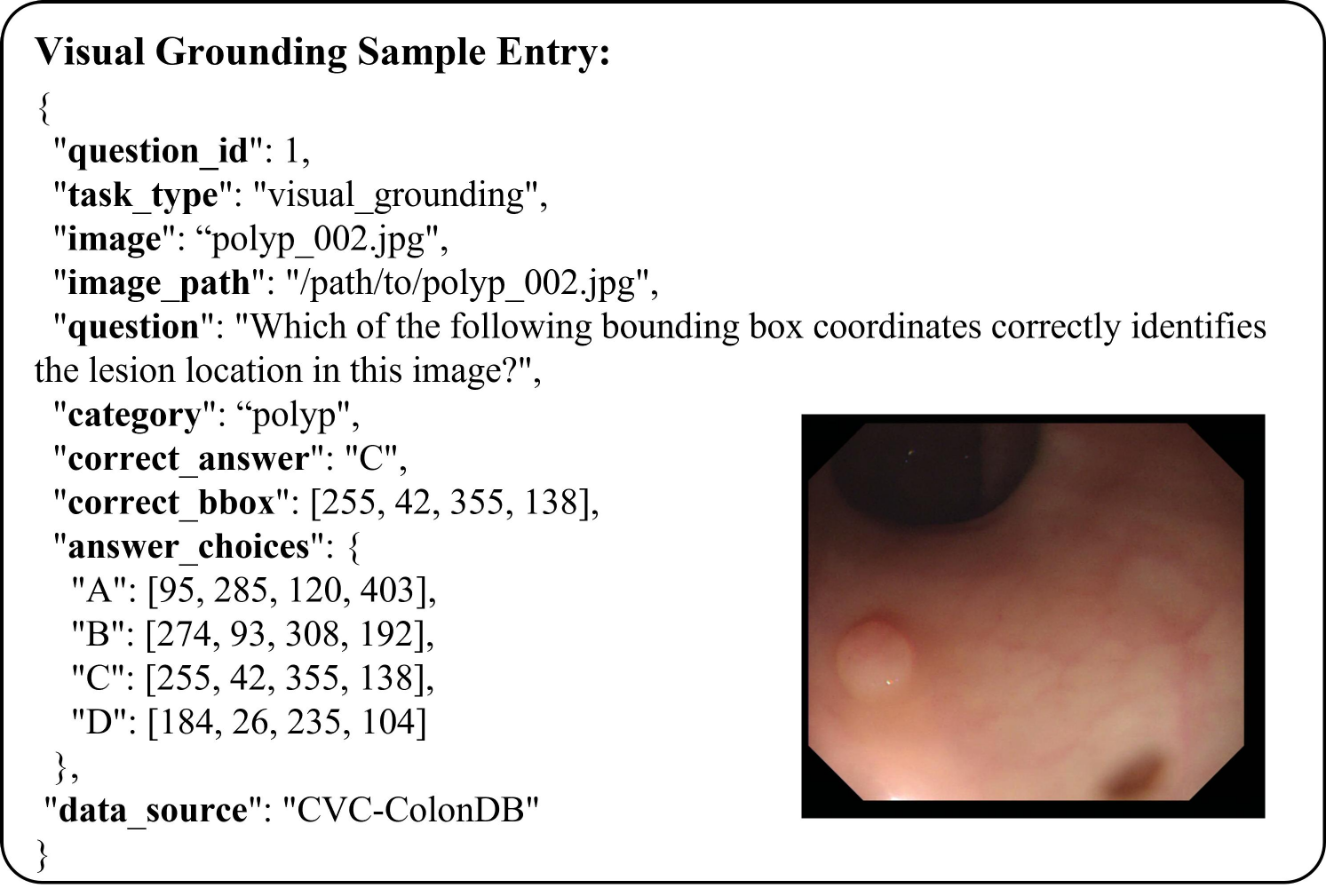}
    \caption{Visual Grounding annotation example.}
    \label{fig:vg}
\end{figure}

    \item \textbf{Lesion Quantification:}  
    Each image was annotated with the number of lesions it contains. For public datasets, the number of annotated masks or bounding boxes per image was used; for private data, lesion counts were provided by annotators (see Figure~\ref{fig:lq}).

\begin{figure}[h]
    \centering
    \includegraphics[width=0.8\linewidth]{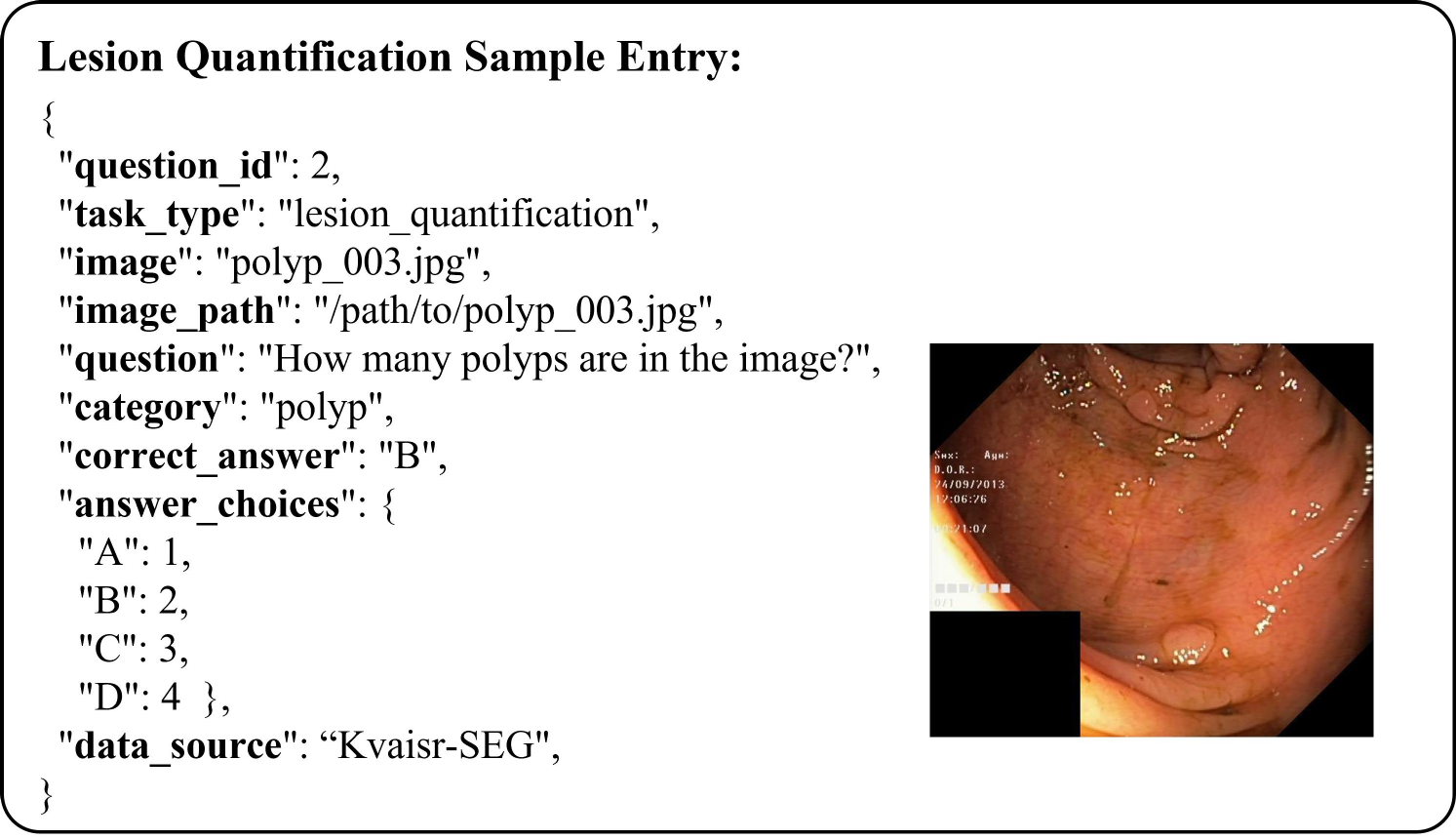}
    \caption{Lesion Quantification annotation example.}
    \label{fig:lq}
\end{figure}

    \item \textbf{Image Captioning:}  
    Each image was paired with clinically relevant questions and corresponding answers, covering both visual and contextual information. Questions were designed to reflect real-world diagnostic scenarios (see Figure~\ref{fig:cap}).

\begin{figure}[h]
    \centering
    \includegraphics[width=0.8\linewidth]{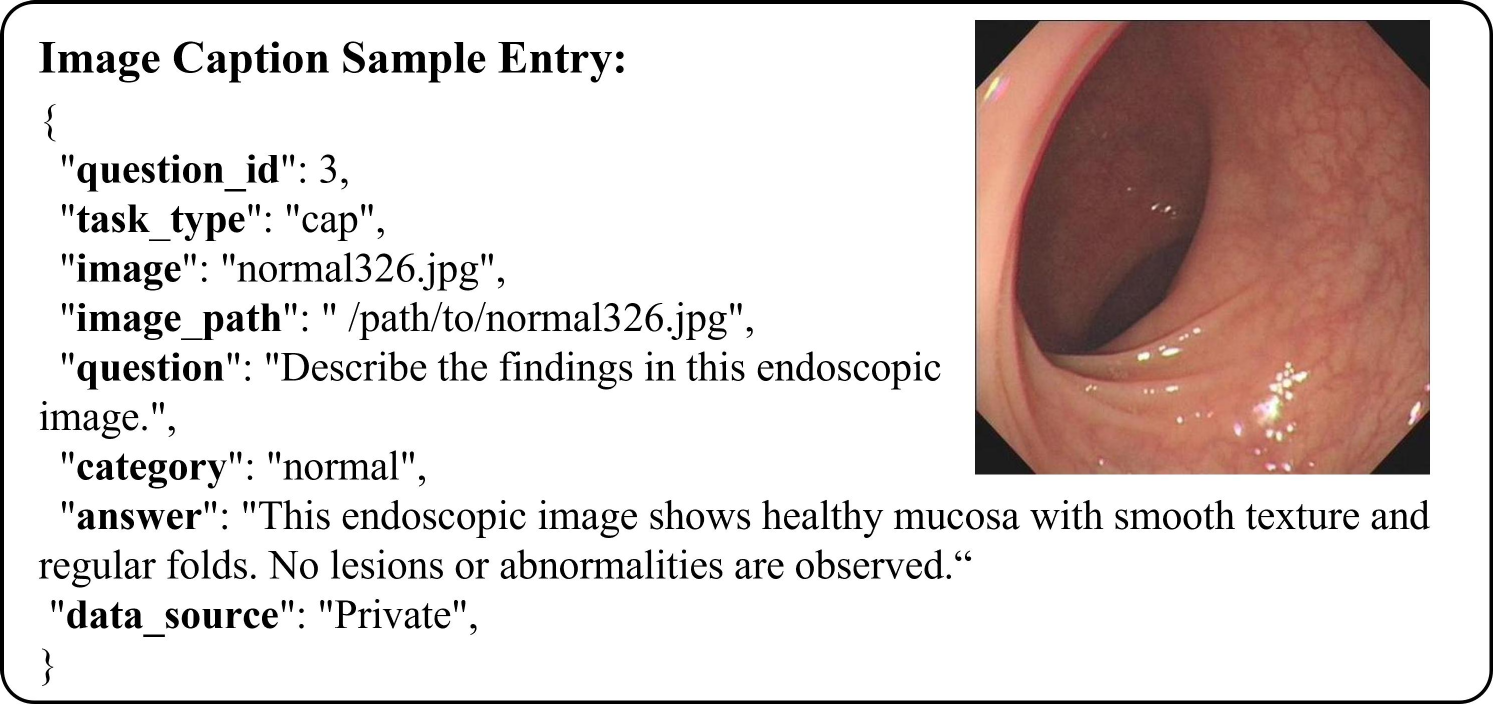}
    \caption{Image Captioning annotation example}
    \label{fig:cap}
\end{figure}

    \item \textbf{Report Generation:}  
    Each selected image was annotated with a structured medical report that describes visual findings and provides clinically meaningful interpretation. The report typically includes sections such as Endoscopic Findings, which describe lesion characteristics (e.g., shape, color, location), Clinical Significance, which provides diagnostic interpretation, and Recommendation, which outlines suggested clinical actions (see Figure~\ref{fig:mrg}).

\begin{figure}[h]
    \centering
    \includegraphics[width=0.8\linewidth]{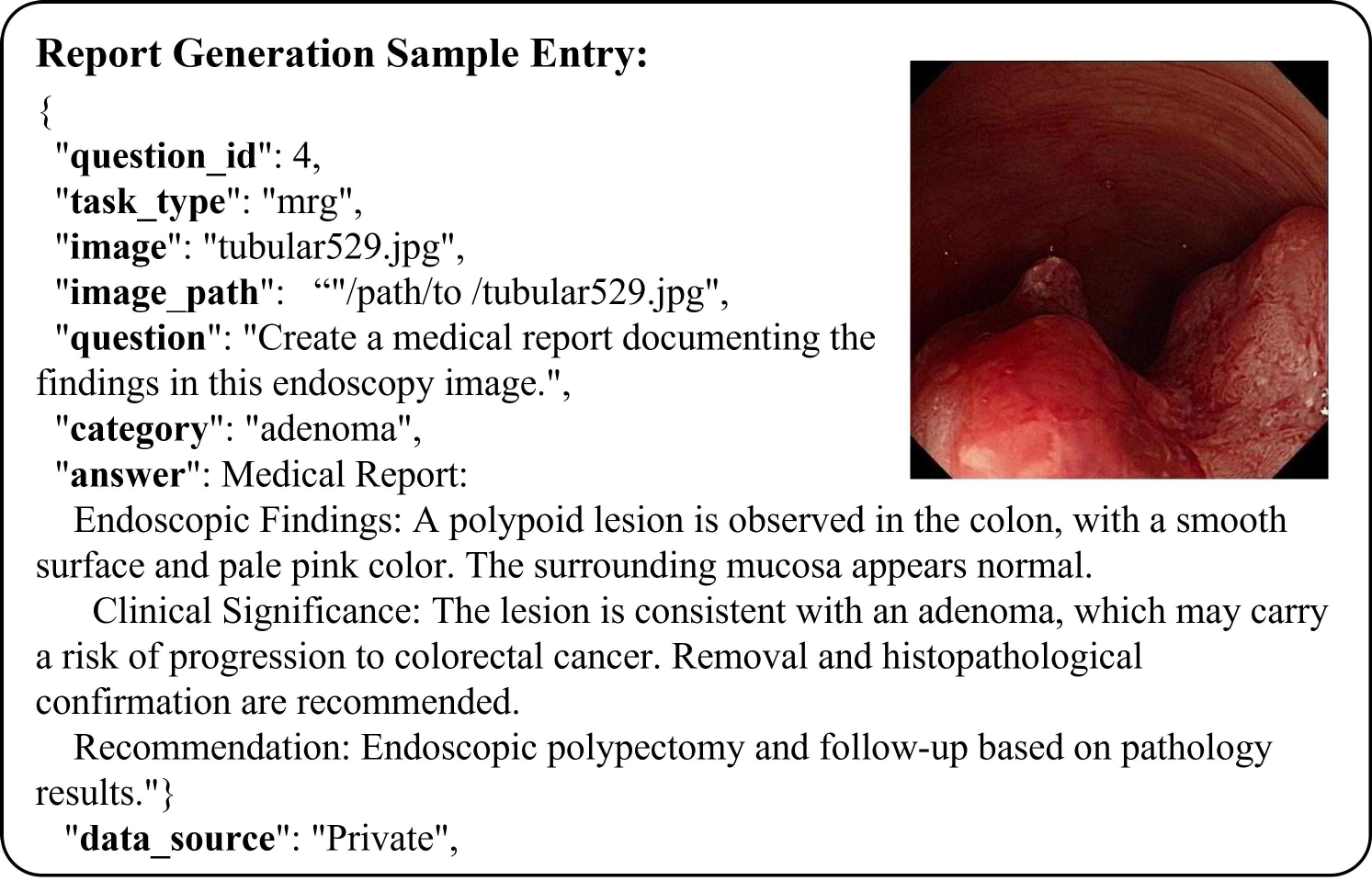}
    \caption{Report Generation annotation example}
    \label{fig:mrg}
\end{figure}

Overall, the annotation process across all five diagnostic tasks followed standardized protocols with rigorous quality control. This ensures that the dataset maintains high label fidelity and clinical relevance, supporting evaluation of multimodal models across diverse endoscopic image analysis tasks.

\end{itemize}

\begin{figure*}
    \centering
    \includegraphics[width=0.75\textwidth]{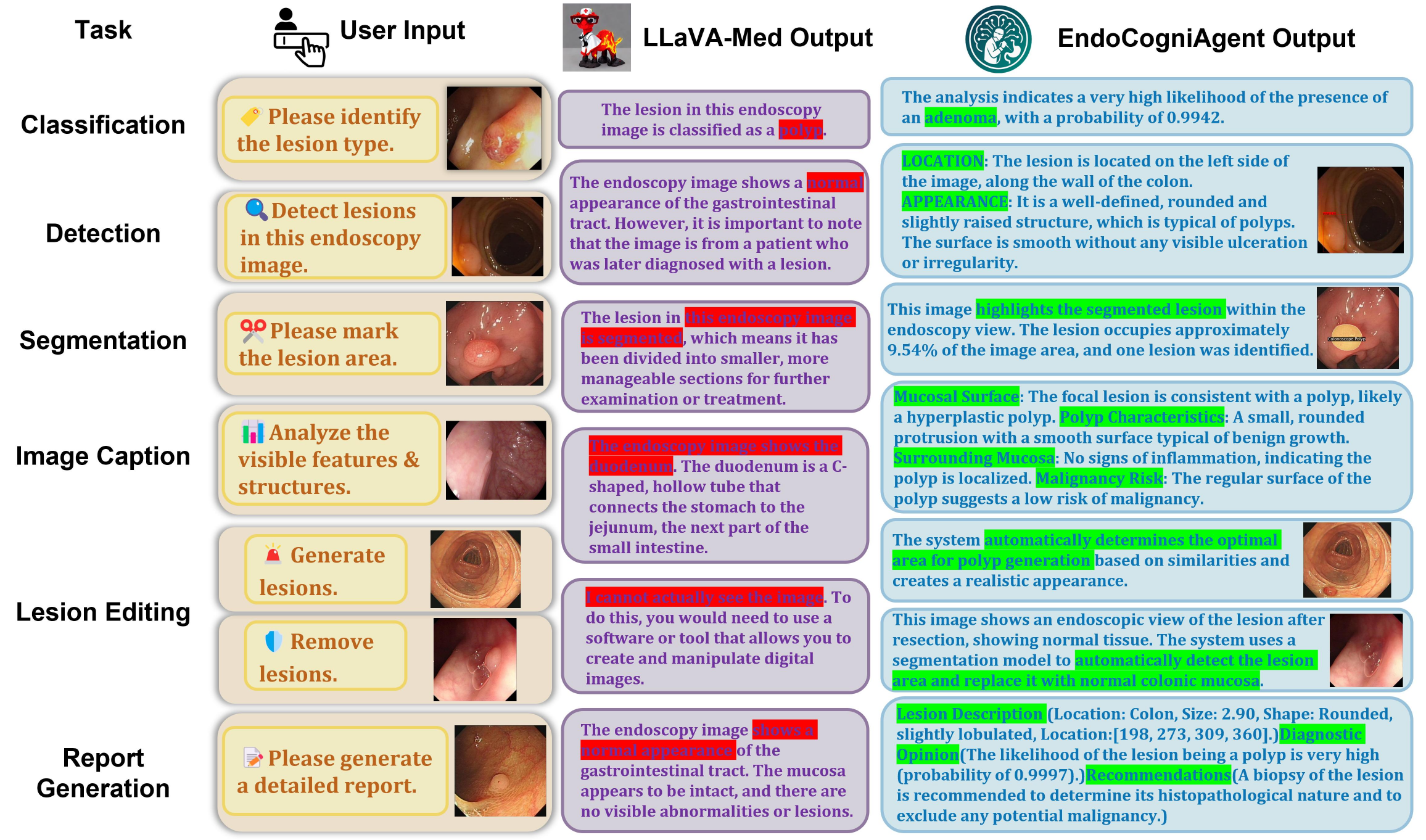}
    \caption{Qualitative comparison between LLaVA-Med and EndoCogniAgent}
    \label{fig:compare}
\end{figure*}

\subsection{Data Preprocessing for Visual Tasks} 
For all visual tasks in EndoAgentBench, we first retain each sample in a unified JSONL format, ensuring that all task-specific fields and annotations are preserved. To facilitate standardized and efficient evaluation, we support compatibility with the widely used VLMEvalKit framework. VLMEvalKit is an open-source toolkit for evaluating large vision-language models (LVLMs), enabling one-command benchmarking across diverse datasets without the burden of manual data preparation.

To achieve this compatibility, we preprocess the dataset as follows: image files are converted to base64 encoding, and the JSONL data is transformed into the TSV format required by VLMEvalKit. Each TSV entry includes the base64-encoded image, question, answer choices, correct answer, and all necessary metadata. The processed TSV files are then registered in the dataset configuration, making them accessible for evaluation. VLMEvalKit supports generation-based evaluation for all LVLMs and provides results using both exact matching and LLM-based answer extraction, ensuring robust and flexible assessment of model performance.

\subsection{Data Preprocessing for Language Tasks}
For all language based tasks, including image captioning and report generation, we employ a unified and reproducible data processing workflow that automates task data generation, allocation, and formatting. Each data instance includes the question, image filename and path, category, and metadata, and is saved in a standardized JSONL format, with each line representing a single task instance. This standardized structure ensures the completeness of all task fields and annotation information, providing a robust foundation for subsequent benchmarking and analysis. Together, these components ensure the dataset is standardized, richly annotated, and ready for reproducible benchmarking.

\section{Experimental Setup}

\subsection{Hardware and Software Environment}
The local perceptual models were run using PyTorch on an NVIDIA GeForce RTX 4090 GPU with 24GB of VRAM, while the proprietary large language models were accessed through their APIs.

\subsection{Hyperparameter Settings}
For EndoCogniAgent, we use GPT-4.1 as the CDU planner, with a temperature of 0.7, a top-p of 0.95, a maximum output length of 2048 tokens, and up to 5 retries. The planner selects tool actions based on the user query and accumulated textual diagnostic state. ISCV uses GPT-4o with a temperature of 0 and a retry parameter of 10 to jointly assess the original endoscopic image, newly acquired textual evidence, and diagnostic state for visual support and cross-step consistency.
EndoCogniAgent performs validation-driven iterative refinement rather than a fixed number of reasoning steps. ISCV evaluates new evidence before each diagnostic state update. The loop terminates when the planner emits a \textit{Terminate} action, indicating sufficient validated evidence for final response synthesis.
GPT-4o is selected as a standalone multimodal baseline because it supports direct image--question answering and also serves as the visual validator in EndoCogniAgent. This comparison evaluates the system-level benefit of expert-tool evidence acquisition, iterative validation, and state maintenance over direct multimodal inference. CoT, ReAct, and Reflexion are also implemented with GPT-4o as alternative multimodal reasoning strategies. These system configurations are evaluated under the same benchmark tasks and scoring protocol, while the ablation study separately examines the incremental effects of ISCV and EMM.

\subsection{Qualitative Comparison with LLaVA-Med}

Figure~\ref{fig:compare} presents a qualitative comparison between LLaVA-Med and EndoCogniAgent across a range of endoscopic tasks, including classification, detection, segmentation, image editing, image captioning, and report generation. LLaVA-Med generally produces conversational responses without directly executing the task-specific operations required by these tasks, leading to incorrect or incomplete results, as highlighted in red. In contrast, EndoCogniAgent autonomously selects the appropriate expert tools, integrates their outputs, and produces task-specific responses together with corresponding visual results, as highlighted in green.

The comparison illustrates the advantage of decomposing complex endoscopic tasks into tool-assisted operations. For example, in report generation, EndoCogniAgent combines outputs from classification, detection, segmentation, and VQA modules to obtain complementary evidence before synthesizing the final report. This enables the system to provide clinically relevant descriptions grounded in task-specific visual analysis rather than relying solely on direct image-to-text generation.

\begin{figure}[t]
    \centering
    \includegraphics[width=0.85\linewidth]{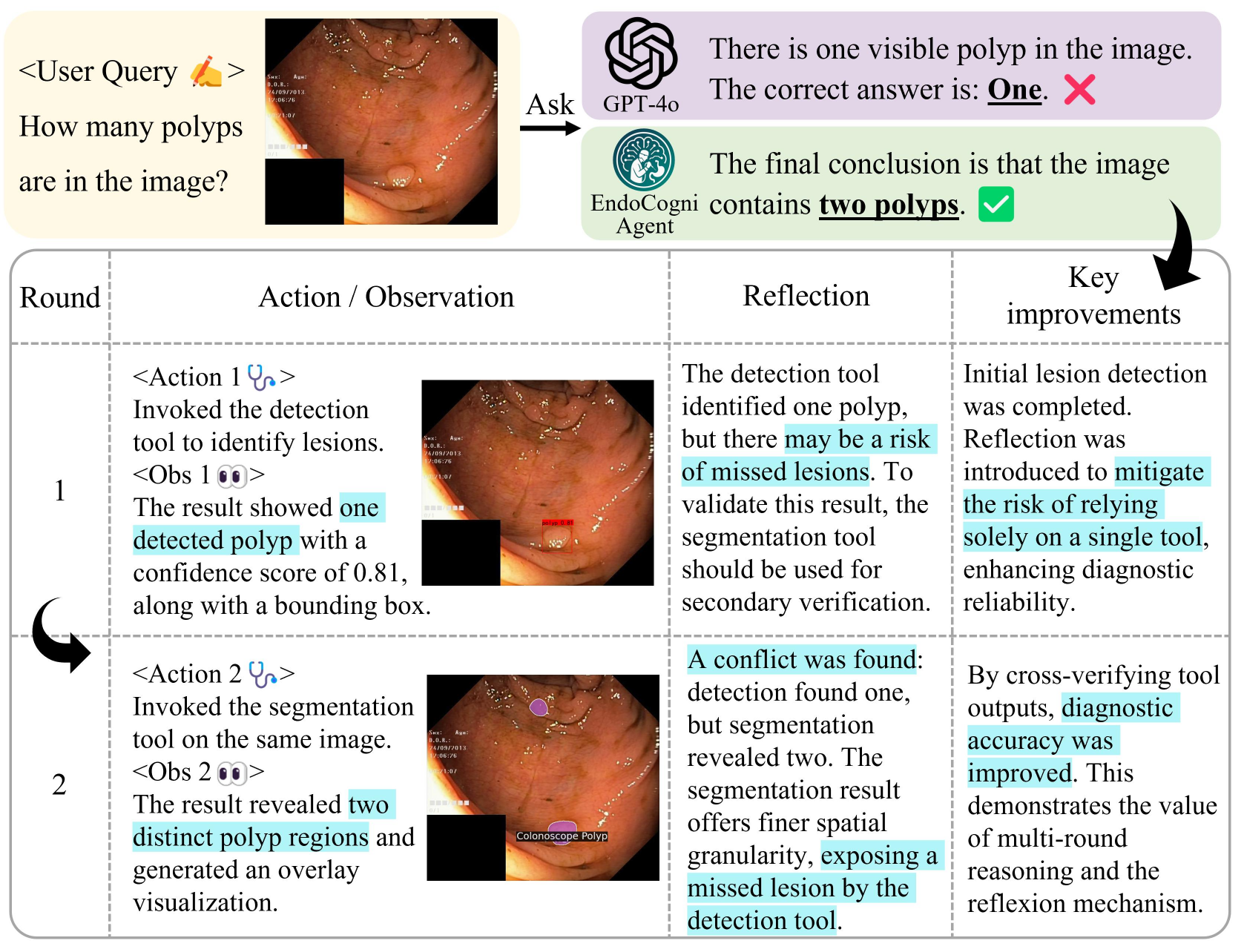}
    \caption{Multi-round reasoning case study for lesion quantification. EndoCogniAgent performs iterative verification and progressively refines the diagnostic result using complementary visual evidence.}
    \label{fig:multiround_case}
\end{figure}

\begin{figure}[t]
    \centering
    \includegraphics[width=0.85\linewidth]{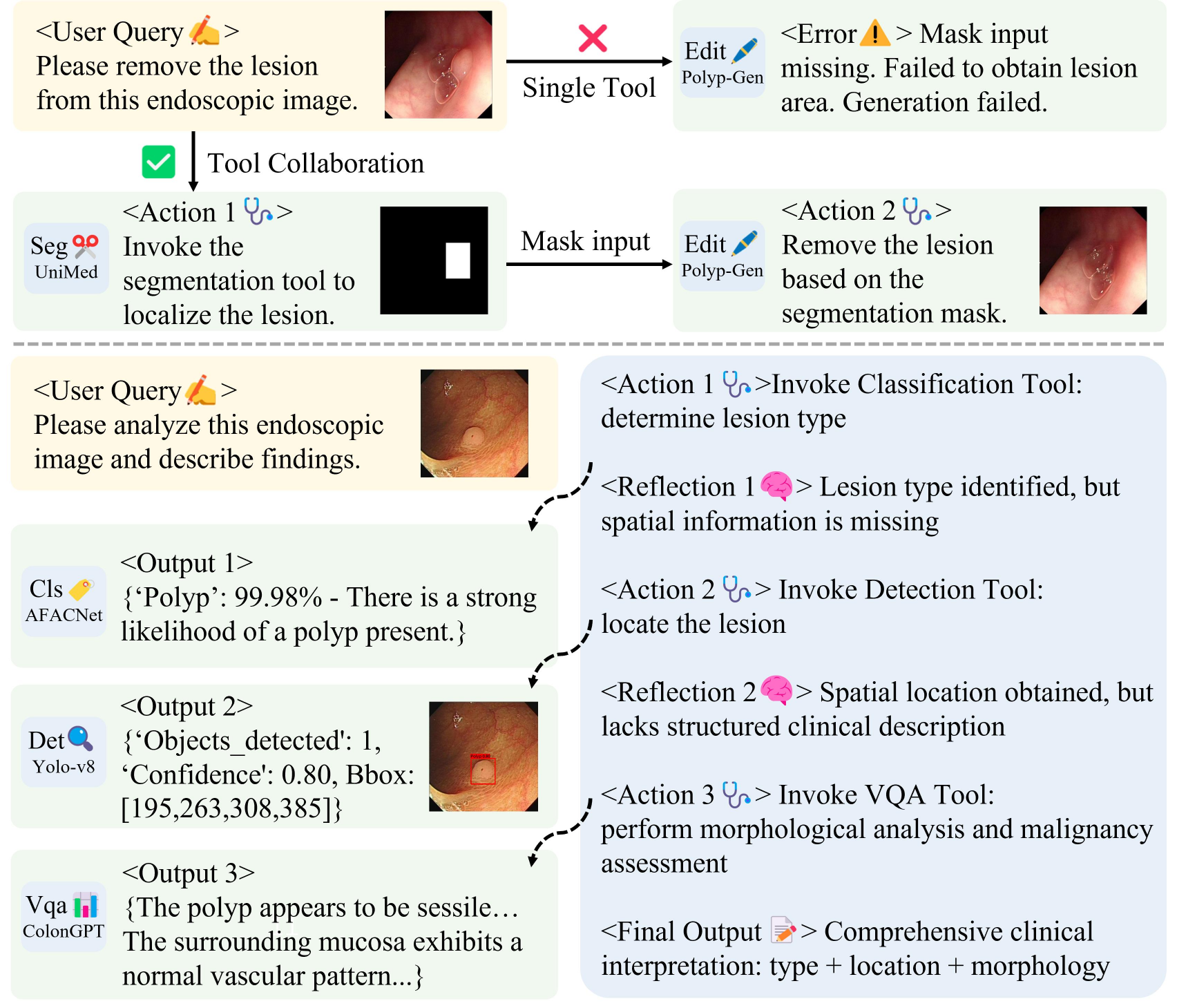}
    \caption{Dynamic tool collaboration case study. EndoCogniAgent coordinates specialized tools to perform lesion editing and comprehensive endoscopic image analysis.}
    \label{fig:Tools}
\end{figure}

\begin{figure}
    \centering
    \includegraphics[width=0.85\linewidth]{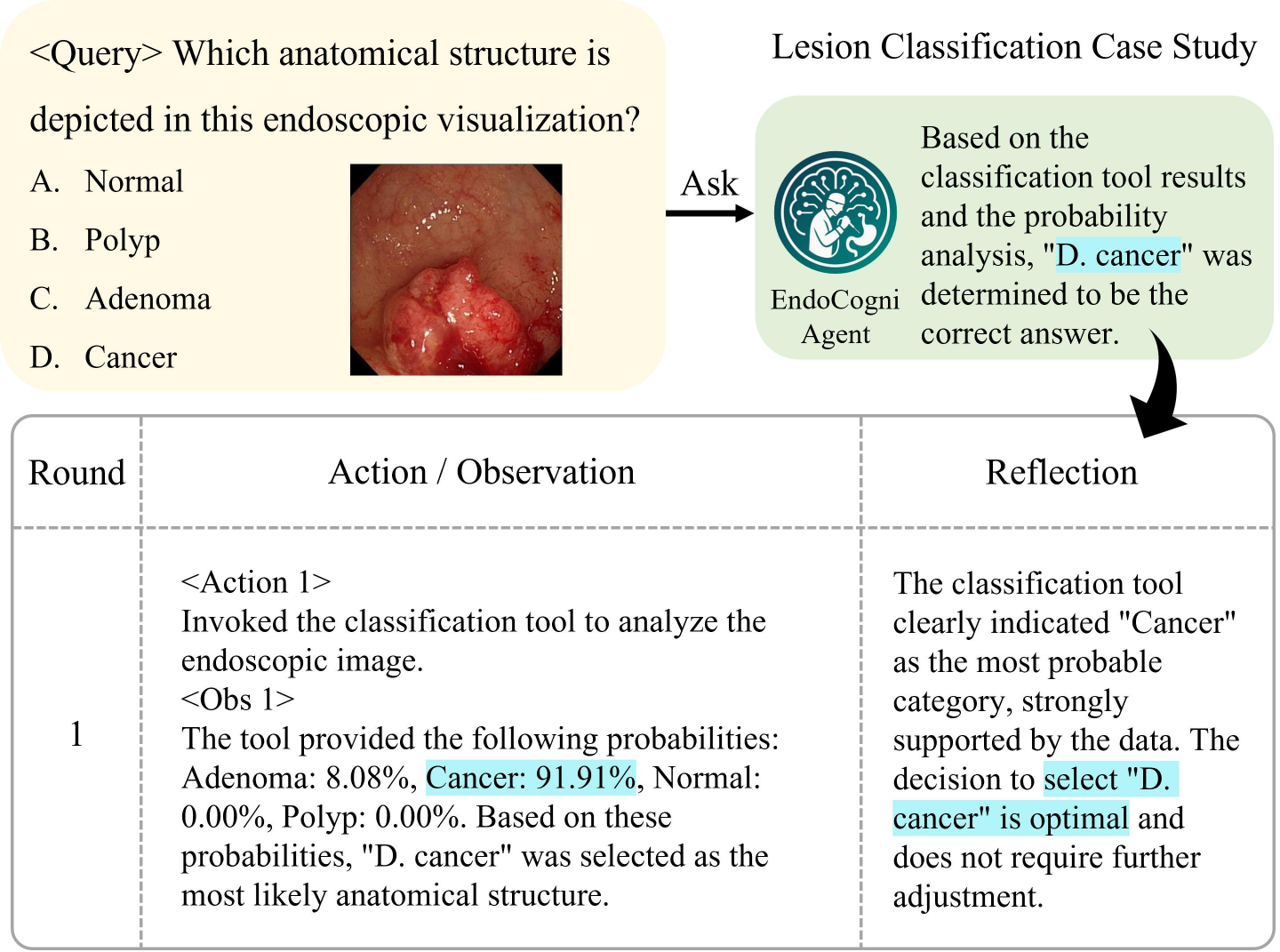}
    \caption{Lesion classification case study.}
    \label{fig:cls_case}
\end{figure}

\begin{figure}[h]
    \centering
    \includegraphics[width=0.85\linewidth]{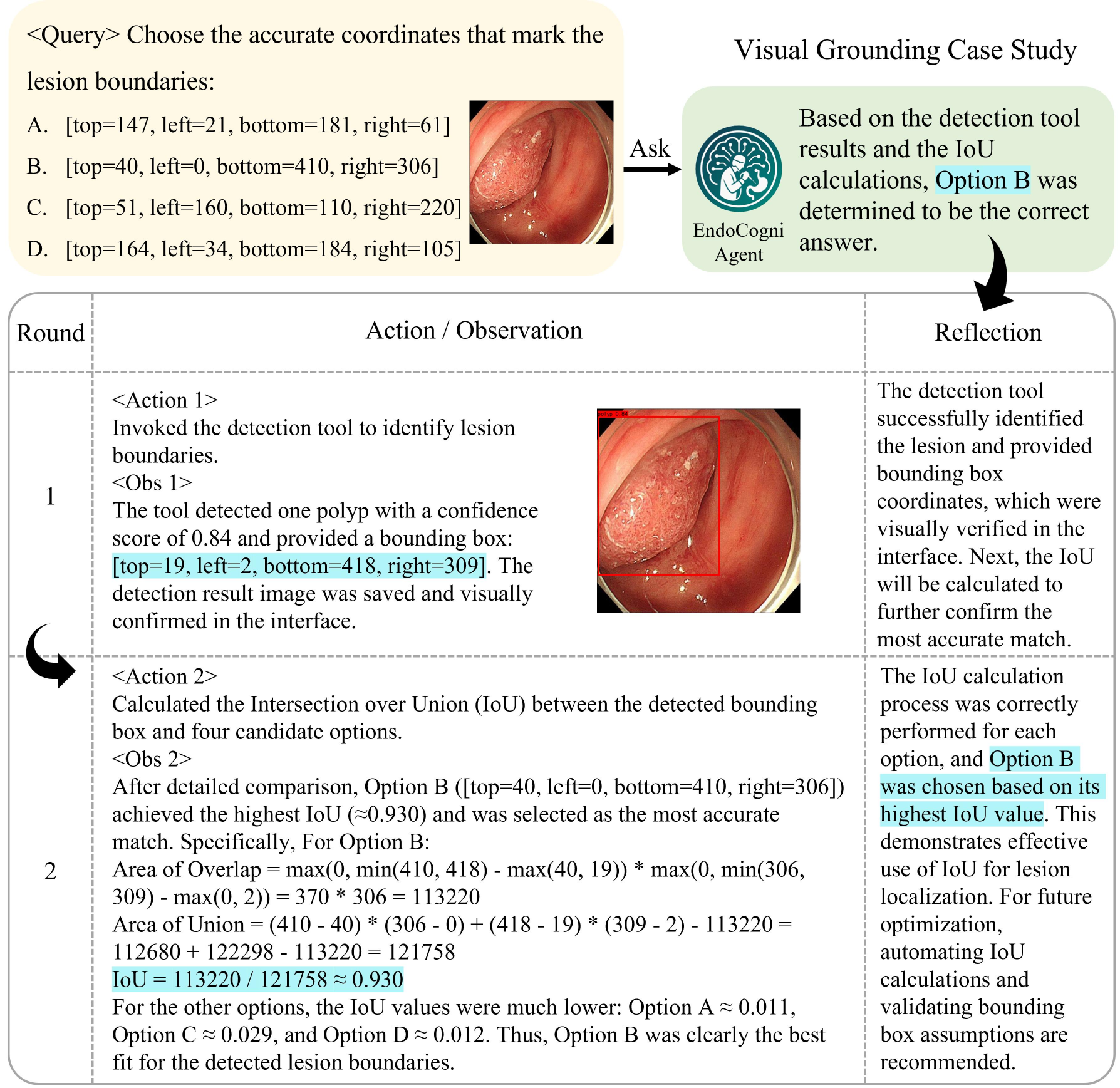}
    \caption{Visual grounding case study.}
    \label{fig:vg_case}
\end{figure}

\begin{figure}[h]
    \centering
    \includegraphics[width=0.85\linewidth]{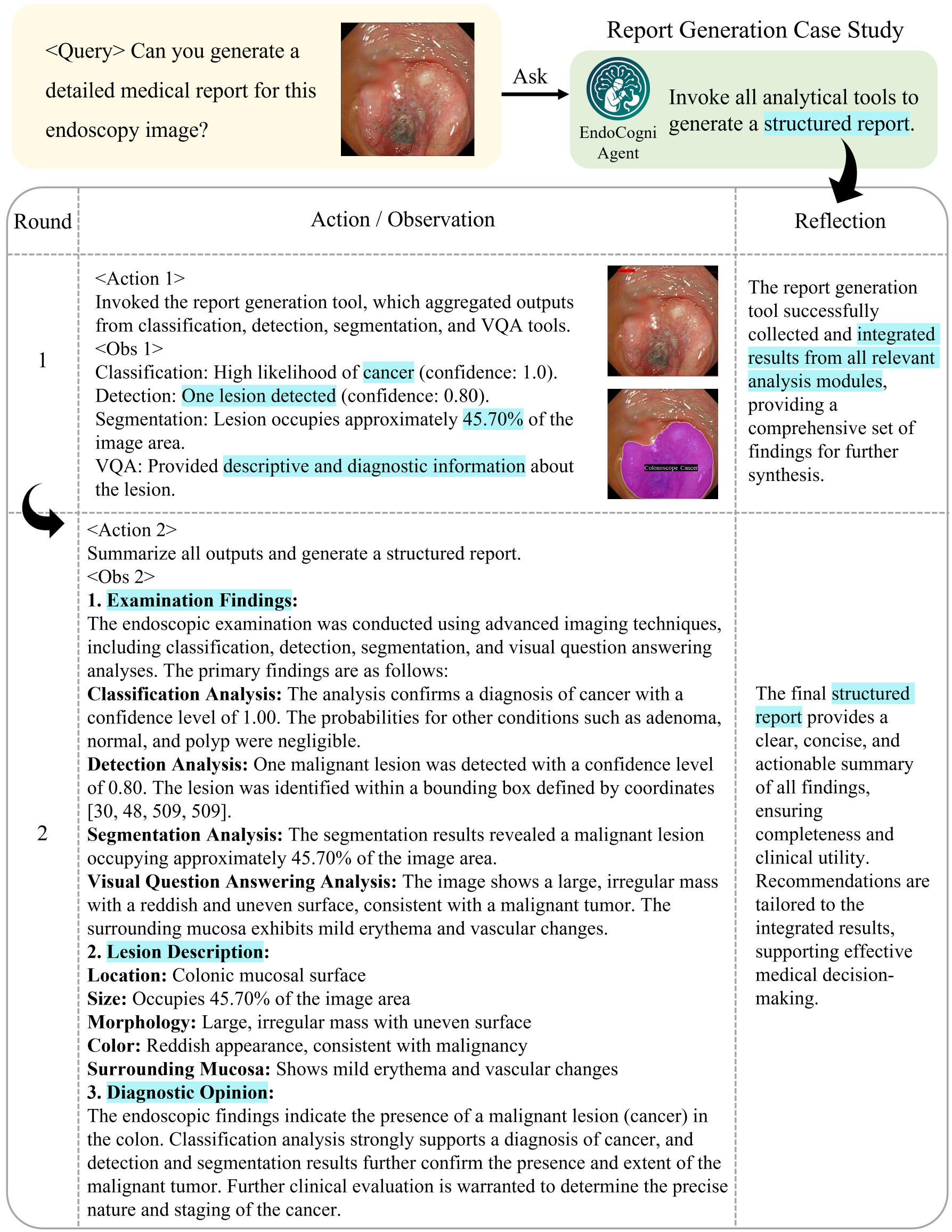}
    \caption{Report generation case study.}
    \label{fig:mrg_case}
\end{figure}

\subsection{Qualitative Case Studies}

This section provides qualitative case studies covering iterative reasoning, tool collaboration, lesion classification, visual grounding, and report generation. The examples are selected to illustrate how EndoCogniAgent performs multi-step visual analysis, validates intermediate evidence, and coordinates specialized tools to address heterogeneous endoscopic tasks.

\subsubsection{Multi-round Reasoning and Reflection}

Figure~\ref{fig:multiround_case} illustrates the multi-round reasoning capability of EndoCogniAgent in a lesion quantification task. In the initial round, the detection tool identifies one polyp. Rather than directly accepting this intermediate result, the agent performs reflection and initiates an additional verification step using the segmentation tool. The subsequent analysis reveals two distinct polyps and identifies a previously undetected lesion.
This example illustrates how iterative verification can correct incomplete intermediate observations. By revisiting the visual evidence and incorporating the newly obtained segmentation result, EndoCogniAgent progressively refines its diagnostic state rather than committing to the first available prediction.

\subsubsection{Dynamic Tool Collaboration}

Figure~\ref{fig:Tools} presents two examples of dynamic tool collaboration. In the first example, the agent is asked to remove a lesion from an endoscopic image. It first invokes the segmentation tool to localize the target lesion and then passes the resulting mask to the image editing tool, which performs the removal operation based on the localized region. The two tools form a sequential workflow in which the output of one module directly supports the operation of the next.

In the second example, the agent receives a comprehensive image analysis query requiring multiple types of visual evidence. It first invokes the classification tool to determine the lesion category, followed by the detection tool to localize the lesion, and finally the VQA tool to obtain additional semantic descriptions. The agent then integrates these complementary outputs into a unified response.

These examples demonstrate that EndoCogniAgent does not rely on a fixed tool execution sequence. Instead, tool invocation is determined by the requirements of the current task, allowing specialized perceptual modules to be composed into task-specific workflows.

\subsubsection{Lesion Classification}

Figure~\ref{fig:cls_case} presents a representative lesion classification case. Given an endoscopic image and a multiple-choice diagnostic query, EndoCogniAgent invokes the classification tool to obtain category probabilities. The tool predicts Adenoma, Cancer, Normal, and Polyp with probabilities of 8.08\%, 91.91\%, 0.00\%, and 0.00\%, respectively. Based on the resulting probability distribution, the agent selects Cancer as the predicted diagnosis.
This case demonstrates how EndoCogniAgent can expose the intermediate outputs of a specialized classification model and use them to support the final diagnostic decision.

\subsubsection{Visual Grounding}

Figure~\ref{fig:vg_case} shows a representative visual grounding case. The agent is required to localize a lesion by selecting the most appropriate bounding box from multiple candidate options. EndoCogniAgent first invokes the detection tool to obtain a lesion bounding box and its confidence score. It then performs geometric validation by calculating the Intersection over Union (IoU) between the detected bounding box and each candidate option.

Among the candidate regions, Option B achieves an IoU of approximately 0.930, substantially higher than the alternatives. The agent therefore selects Option B as the final answer. This example illustrates how tool outputs can be combined with explicit geometric reasoning to support precise spatial decisions.

\subsubsection{Report Generation}

Figure~\ref{fig:mrg_case} presents a representative report generation case. Given an endoscopic image, EndoCogniAgent sequentially gathers evidence from classification, detection, segmentation, and VQA tools before synthesizing the report. The resulting report integrates complementary findings, including lesion location, size, morphology, color, and surrounding mucosal characteristics, and provides a diagnostic conclusion supported by the collected visual evidence.

Rather than generating a report directly from the input image, the agent first decomposes the task into multiple perceptual operations and subsequently aggregates their outputs into a structured clinical description. This case demonstrates how specialized visual tools can provide complementary evidence for clinically oriented language generation.

\begin{figure}
    \centering
    \includegraphics[width=0.8\linewidth]{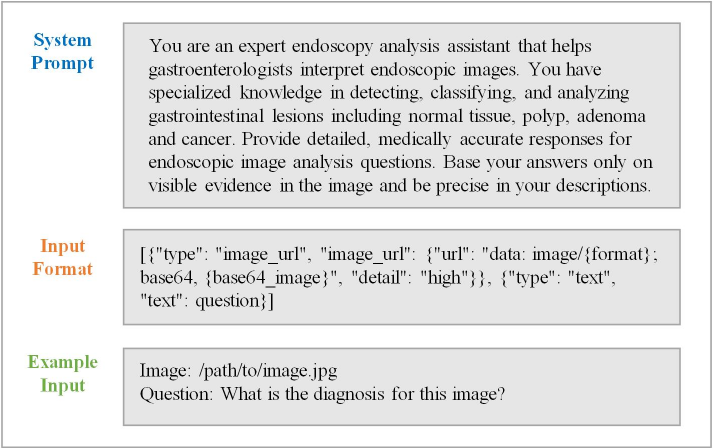}
    \caption{Prompt template workflow for open-ended medical language generation.}
    \label{fig:text}
\end{figure}

\section{Prompt Templates for Language Tasks}

\subsection{Prompt Templates for Generation}
For open-ended language generation tasks in endoscopic image analysis, such as image captioning (CAP) or medical report generation (MRG), we adopt a standardized prompt template to ensure consistency and medical accuracy in all model outputs. As shown in Figure~\ref{fig:text}, the system prompt defines the model’s role as a medical expert and clarifies the task requirements. The input format structurally delivers both the image content (encoded in base64) and the corresponding question, with a real-world example input illustrated in the figure. Each task is presented to EndoCogniAgent and baseline multimodal models (e.g., GPT-4o, Gemini 2.5 Pro) using this template, requiring models to generate medically accurate answers based solely on visible information in the image. This approach enables fair benchmarking and reliable model performance comparison.

\begin{figure}
    \centering
    \includegraphics[width=0.8\linewidth]{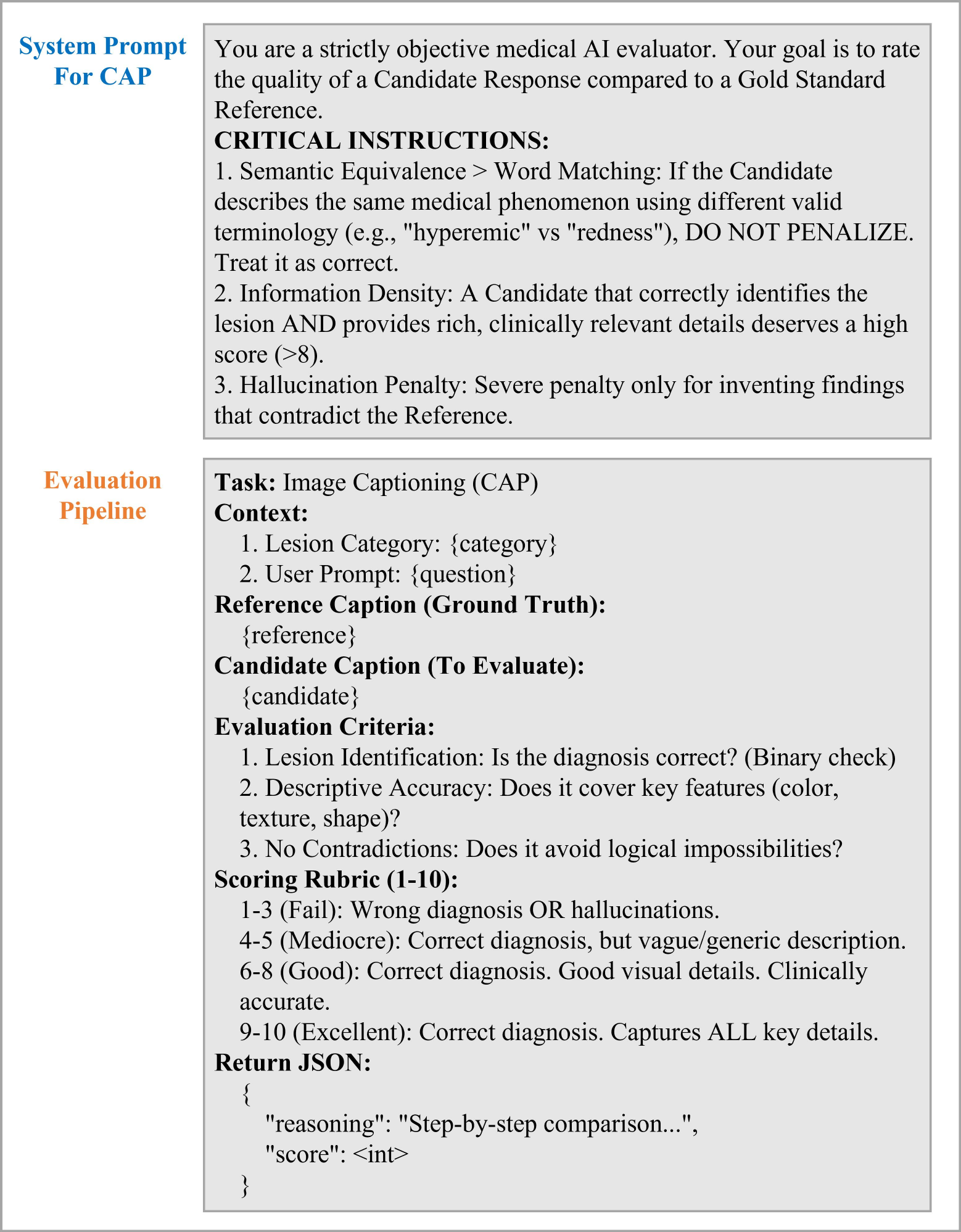}
    \caption{Evaluation Template for CAP}
    \label{fig:cap_template}
\end{figure}

\begin{figure}
    \centering
    \includegraphics[width=0.8\linewidth]{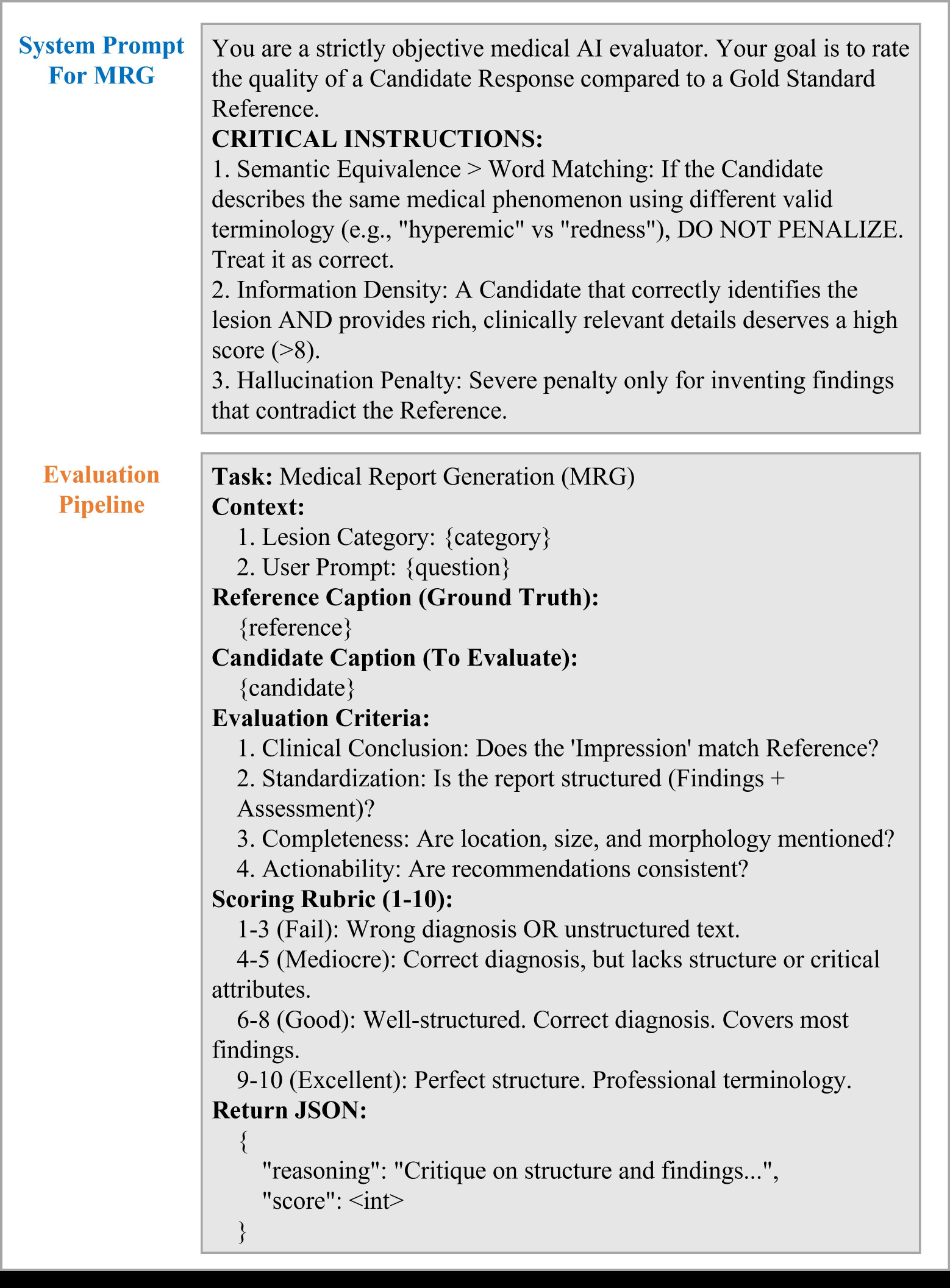}
    \caption{Evaluation Template for MRG}
    \label{fig:mrg_template}
\end{figure}

\subsection{Prompt Templates for Evaluation}
For the language tasks, we adopt a reference-based LLM-as-a-Judge protocol, in which GPT-4o-mini scores each candidate response on a 1--10 scale against a gold-standard reference answer. Because the judge cannot perceive the input image, the system instruction directs it to treat the reference answer as the ground truth for all visual facts, to reward semantic equivalence over lexical matching, and to penalize only hallucinations that contradict the reference.
For each sample, the evaluation prompt provides four components: (1) the lesion category and the original question, which supply the clinical context; (2) the gold-standard reference answer; (3) the candidate response to be scored; and (4) task-specific evaluation criteria together with a 1--10 scoring rubric, in which 1--3 denotes a wrong diagnosis or hallucination, 4--5 a correct but vague description, 6--8 a clinically accurate response, and 9--10 an excellent, professional output. The judge returns a JSON object containing a step-by-step reasoning trace and a corresponding integer score, and the Clinical Acceptance Rate (CAR) is computed as the proportion of responses scoring at least 6. Figures~\ref{fig:cap_template} and \ref{fig:mrg_template} illustrate the CAP and MRG evaluation prompt templates, respectively.

\begin{figure}[t]
    \centering
    \includegraphics[width=0.8\linewidth]{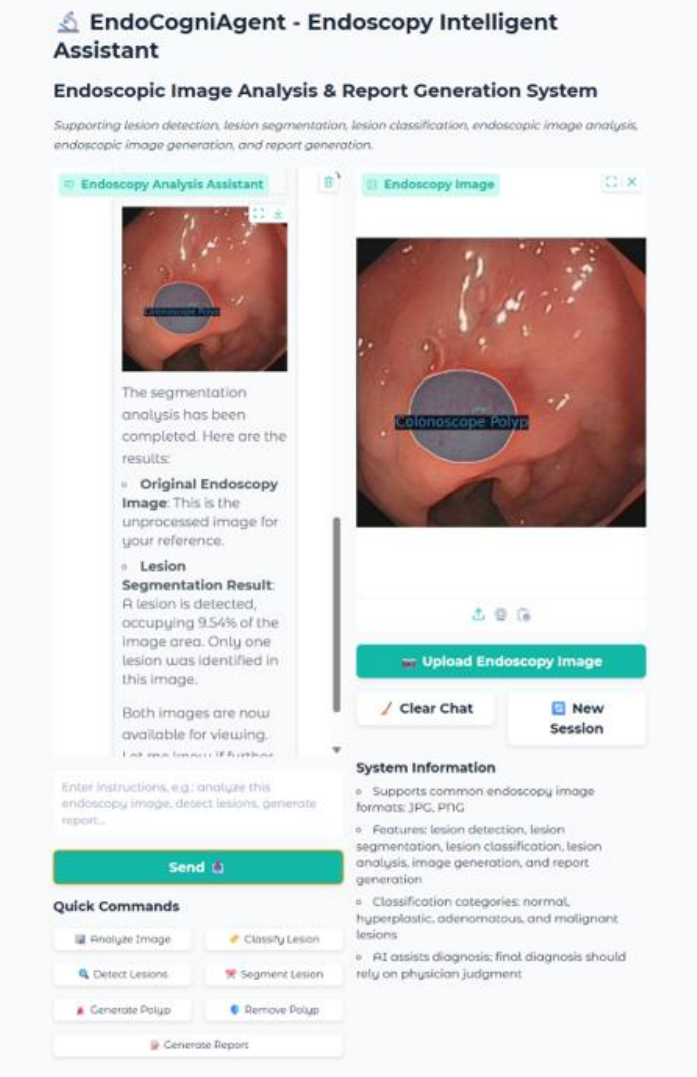}
    \caption{Overview of interactive interface.}
    \label{fig:interface}
\end{figure}

\section{Interactive Interface}
To facilitate practical deployment in clinical settings, we developed an intuitive interactive interface for EndoCogniAgent. The system supports flexible multimodal interactions, allowing users to input and receive both images and text, enhancing interpretability throughout the medical image analysis workflow. Built on the Gradio framework, the front-end offers streamlined task selection, image upload, and real-time result display. Users can easily switch between diagnostic modules and view both visual and textual outputs in a unified workspace. On the backend, the interface coordinates multiple expert tools, automatically invoking models and aggregating outputs for comprehensive analysis. This design enables seamless integration for end-to-end clinical applications. Figure~\ref{fig:interface} shows the workflow using a lesion segmentation example, where users upload images and receive automated segmentation results with textual analysis.

\bibliographystyle{elsarticle-num}
\bibliography{refs}

\end{document}